\documentclass[10pt, letterpaper]{article}
\usepackage{graphicx}
\usepackage{color}
\usepackage{epsfig}
\usepackage{amsmath}
\usepackage{amssymb}
\usepackage{amsthm}
\usepackage{subfigure}

\makeatletter
\renewcommand{\@thesubfigure}{}

\newtheorem{Property}{Property}

\newtheorem{Corollary}{Corollary}

\providecommand{\myproofname}{\footnotesize{Proof}}

\begin{document}

\title{Inner Product Similarity Search using Compositional Codes} 



\author{Chao Du$^{1}$~~~~~Jingdong Wang$^{2}$ \\
$^{1}$Tsinghua University, Beijing, P.R. China\\$^{2}$Microsoft Research, Beijing, P.R. China}

\maketitle

\begin{abstract}
This paper addresses the nearest neighbor search
problem
under inner product similarity
and introduces a compact code-based approach.
The idea is
to approximate a vector
using the composition of several elements selected from a source dictionary
and to represent this vector
by a short code composed of
the indices of the selected elements.
The inner product between a query vector
and a database vector
is efficiently estimated
from the query vector and the short code of the database vector.
We show the superior performance of the proposed group $M$-selection algorithm
that selects $M$ elements from $M$ source dictionaries for vector approximation
in terms of search accuracy and efficiency
for compact codes of the same length
via theoretical and empirical analysis.
Experimental results
on large-scale datasets ($1M$ and $1B$ SIFT features,
$1M$ linear models and Netflix)
demonstrate the superiority of the proposed approach.
\end{abstract}

\section{Introduction}
Similarity search~\cite{SameFoun2006} is a fundamental research topic
in the area of
computational geometry and machine learning.
It has attracted a lot of interests~\cite{ShakhnarovichDI06}
in computer vision and pattern recognition
because of
the popularity of large scale and high-dimensional multimedia data.
Various technologies,
such as index structures~\cite{Bentley75,MujaL09} and compact codes~\cite{DatarIIM04,JegouDS11},
have been developed
to solve the similarity search problem
under different similarity metrics~\cite{Andoni09}.

In this paper, we are interested in
designing a compact code approach
with a focus on inner product similarity.
Inner product similarity search is an important task
in many vision applications.
In large scale retrieval of images from text queries
and multi-class categorization,
the PAMIR (Passive-Aggressive Model for Image
Retrieval) approach~\cite{GrangierB08}
trains a large number of linear models
(each corresponds to a text query),
and ranks the queries for a new image
according to the scores
evaluated over the linear models,
which is an inner product similarity search problem.
In the object detection task
with a large number of object classes~\cite{DeanRSSVY13},
it consists of a step finding the top-responded filters
by performing the convolution operation over sliding image windows
and filters,
which is also an inner product similarity search problem.
The latent factor models
widely used in recommendation systems~\cite{KorenBV09}
and document matching~\cite{BayardoMS07},
such as matrix factorization~\cite{KorenBV09},
latent semantic index~\cite{DeerwesterDLFH90},
and so on,
also rely on inner product similarity search
to find the best matches.

We propose a compact code approach
to approximate inner product similarity search.
Our approach is based on a vector approximation algorithm,
using the composition of several vectors selected
from a small set (source dictionary) as the approximation,
which is not studied for compact codes and similarity search before.
Then we use the indices of the selected vectors
to form a compact code,
which we call compositional codes,
to describe the data vector.
Finally, inner product between the query vector and the database vector
can be efficiently estimated
from the query vector and the code of the database vector.

The compositional way to vector approximation
can be viewed as a quantization algorithm,
finding the nearest element
from a larger dictionary (called compositional dictionary)
that is produced
from a source dictionary.
We study the way of
using $M$-selection ($M$-combination with repetitions)\footnote{In mathematics,
an $M$-combination of a set $\mathcal{S}$ is a subset of $M$ distinct elements of $\mathcal{S}$.
An $M$-selection of a set $\mathcal{S}$ is a subset of $M$ not necessarily distinct elements of $\mathcal{S}$.}
to form the compositional dictionary,
and show that it is equivalent to
performing $1$-selection
from $M$ identical source dictionaries respectively.
This equivalence motivates us to generalize $M$-selection
by using $M$ different source dictionaries,
yielding a so-called group $M$-selection algorithm
that simultaneously learns source dictionaries
and performs joint $M$ $1$-selections from source dictionaries.
The advantage of group $M$-selection lies in
more accurate vector approximation
because of a larger compositional dictionary
but with the compact code of the same length.
Experimental results on finding similar SIFT features,
searching for users with similar interests
and discovering most relevant liner models
show excellent search accuracy.

\subsection{Related Work}
Similarity search (or nearest neighbor search)
has been studied in many research areas,
including computational geometry,
computer vision,
machine learning,
data mining
and so on.
A lot of algorithms have been developed
for approximate nearest neighbor search,
under
the Euclidean distance~\cite{Bentley75,SameFoun2006},
the earth mover distance~\cite{Andoni09,GraumanD04},
and so on.
In this paper,
we are interested in the nearest neighbor search problem,
instead under the inner product similarity.

The challenges, compared with the well-studied
similarity search under Euclidean distance, are analyzed
in~\cite{RamG12}. The main difficulty that inner product does
not satisfy the triangle equality makes algorithms depending on
it not suitable for inner product search. A cone tree based
index structure~\cite{RamG12} is designed for exact inner product
similarity search. The fact that exact search under Euclidean
distance in high-dimensional cases is even slower than the
naive linear scan algorithm is also observed under inner product
similarity. Thus, we focus on approximate
inner product similarity search and study the compact code
approach.

There are many algorithms based on compact codes
for similarity search with the Euclidean distance,
including two main categories:
hashing and compression-based source coding.
The hashing category consists of
random algorithms,
such as locality sensitive hashing~\cite{DatarIIM04},
and learning based algorithms
such as spectral hashing~\cite{WeissTF08},
iterative quantization~\cite{GongL11}
and so on.
These algorithms show promising performance for searching with Euclidean distance,
but most of them cannot be directly applied for inner product similarity.
The compression-based source coding category includes
$k$-means, product quantization~\cite{JegouDS11},
and Cartesian $k$-means~\cite{NorouziF13},
which are shown to achieve superior
performance over hashing codes
with a little higher but still acceptable query time cost.
The proposed approach belongs to the compression-based category,
with a specific adaptation to inner product.
The closely related approaches,
product quantization
and Cartesian $k$-means,
we will show,
are constrained versions of
our proposed approach.

Recent research on hyperplane hashing~\cite{JainVG10,LiuWMKC12,MuWC12}
studies the problem of finding the points
that are nearest to the hyperplane, which is related to inner
product. Different from the maximum inner product problem our
approach addresses, it is equivalent to finding the data vector
that has the minimum inner product with the query vector.
Concomitant hashing~\cite{MuWC12} is also able to solve the absolute
maximum inner product problem.
Approximate nearest subspace search~\cite{BasriHZ11}, under the
similarity based on the principal angles between subspaces, is
related to the cosine similarity search. Those approaches
address different problems and are not comparable to our
approach.

\section{Inner Product Similarity Search}\label{section:InnerProductSimilaritySearch}
Given a set of $N$ $d$-dimensional  database vectors
$\mathcal{X} = \{\mathbf{x}_1, \mathbf{x}_2, \cdots, \mathbf{x}_N\}$
and a query $\mathbf{q}$,
inner product similarity search aims to find a database vector
$\mathbf{x}^*$
so that $\mathbf{x}^* = \arg\max_{\mathbf{x} \in \mathcal{X}} \texttt{<} \mathbf{q}, \mathbf{x}\texttt{>}$.

In this paper,
we study the approximate inner product similarity search problem
with a focus on the compact coding approach,
i.e., finding short codes
to represent the database vectors.
The objective includes three aspects:
the code representing the database vector is compact;
the similarity between a query $\mathbf{q}$
and a vector
can be accurately approximated
using the query and the compact code;
and the evaluation over the query and the compact code
can be quickly conducted.

The basic idea of our approach is to approximate a database vector $\mathbf{x}_n$
using a compositional vector,
the summation $\sum_{m=1}^M \mathbf{c}_{n_m}$
of $M$ exemplar vectors
$\{\mathbf{c}_{n_1}, \mathbf{c}_{n_2}, \cdots, \mathbf{c}_{n_M}\}$,
where the exemplar vectors are selected
from a collection of exemplars
$\mathcal{C}$.
The main work of this paper is to investigate its application
to compact codes and effective and efficient inner product similarity approximation.
Suppose that each example in $\mathcal{C}$ can be
represented by a code of length $\log K$,
where $K$ is the size of $\mathcal{C}$.
Then the compositional vector
and thus the database vector
can be represented
by a short code of length $M\log K$.

The proposed approach
also
exploits the distributive property
with respect to the inner product operation ($\texttt{<}\cdot, \cdot\texttt{>}$)
over the addition operation:
$\texttt{<}\mathbf{q}, \mathbf{c}_1 + \mathbf{c}_2\texttt{>} = \texttt{<}\mathbf{q}, \mathbf{c}_1\texttt{>}
+ \texttt{<}\mathbf{q}, \mathbf{c}_2\texttt{>}$.
With the distribution property,
evaluating inner product
between a query and the compositional vector
takes $O(M)$ addition operation
if the inner product values of $\mathbf{q}$ with all vectors
in $\mathcal{C}$ are computed,
whose time cost is
neglectable when handling large scale data.

The vector approximation scheme using $M$ vectors
is expected to have a better approximation,
thus yielding a more accurate inner product approximation.
This is guaranteed by
the property
that the inner product approximation error
is upper-bounded
if vector approximation is
with an upper-bounded error
(Euclidean distance with vector approximation
has a similar property derived
from the triangle inequality).
\begin{Property}
\label{property:innerproductbound}
Given a data vector $\mathbf{p}$
and a query vector $\mathbf{q}$,
if the distance between $\mathbf{p}$
and its approximation $\bar{\mathbf{p}}$
is not larger than $r$,
$\|\mathbf{p} - \bar{\mathbf{p}}\|_2 \leqslant r$,
then the absolute difference between
the true inner product
and the approximate inner product
is upper-bounded:
\begin{align}
|\texttt{<} \mathbf{q}, \mathbf{p} \texttt{>} - \texttt{<} \mathbf{q}, \bar{\mathbf{p}} \texttt{>}|
\leqslant r \|\mathbf{q}\|_2.
\end{align}
\end{Property}

The upper bound $r\| \mathbf{q} \|_2$
is related to the $L_2$ norms of $\mathbf{q}$,
meaning that the bound depends on the query $\mathbf{q}$
(in contrast,
the upper bound for Euclidean distance does not depend on the query).
However,
the solution in inner product similarity search
does not depend on the $L_2$ norm of the query
as queries with different $L_2$ norm have the same solution,
i.e.,
$\mathbf{x}^* = \arg\max_{\mathbf{x} \in \mathcal{X}} \texttt{<} \mathbf{q}, \mathbf{x}\texttt{>}
= \arg\max_{\mathbf{x} \in \mathcal{X}} \texttt{<} s\mathbf{q}, \mathbf{x}\texttt{>}$,
where $s$ is an arbitrary positive number.
In this sense, it also holds
that
more accurate vector approximation is potential to
lead to better inner product similarity search.

\section{Our Approach}
In this section,
we first introduce the basic vector approximation approach,
$k$-means clustering,
and connect it with the manner of
using a $1$-combination of a source dictionary
to approximate the data vector.
We then present the proposed compositional code approach,
based on $M$-combination, $M$-selection and group $M$-selection.
Finally,
we give the analysis.

\subsection{$K$-means}
$K$-means clustering is a method of vector quantization.
It aims to partition the database points
into $K$ clusters
whose centers form a set $\mathcal{C}$,
in which each database point belongs to the cluster with the nearest center.
In its application to data approximation,
each database point is approximated
by the nearest center,
equivalently using the best $1$-combination
of $\mathcal{C}$
to approximate a database vector.
The $K$-means clustering algorithm
provides a way of jointly optimizing the center set
$\mathcal{C}$ and the $1$-combination for each data vector.


\subsection{Compositional Codes}\label{subsection:OurApproachCompositionalCodes}
We present the basic compositional code approach
that uses $M$-combination over a set of samples
denoted by $\mathcal{C}$,
i.e., using the compositional $\sum_{m=1}^M \mathbf{c}_{n_m}$
with $\mathbf{c}_{n_m} \in \mathcal{C}$,
to approximate the data vector $\mathbf{x}_n$.
This manner could be viewed as a two-step scheme:
first producing a \emph{compositional} dictionary
formed by the $M$-combinations of
$\mathcal{C}$ that we call \emph{source} dictionary
and then finding the best element from the composite dictionary
as the approximation of a data vector.

Instead of separately learning the source dictionary
(e.g., using $k$-means clustering)
and finding the optimal $M$-combinations,
we use the way similar to sparse coding and $k$-means
to jointly learn the source dictionary and $M$-combinations.
We use a $K$-dimensional binary vector $\mathbf{b}_n$
to represent an $M$-combination,
where only $M$ entries in $\mathbf{b}$ are valued as $1$
and all others are $0$,
and a matrix $\mathbf{C}$
of size $d \times K$
to represent the source dictionary
with each column corresponding to an item
of the source dictionary.
The objective function is written as follows,
\begin{align}
\min_{\mathbf{C}, \mathbf{b}_1, \cdots, \mathbf{b}_N} \sum\nolimits_{n=1}^N \|\mathbf{x}_n - \mathbf{C}\mathbf{b}_n\|_2^2.\label{eqn:compositecoding}
\end{align}

We relax the constraint in $M$-combination
that the $M$ elements in $M$-combination are distinct
and use $M$-selection in which the elements are not necessarily different.
To mathematically formulate this case,
we use a longer binary vector $\mathbf{b}_n$
whose dimensionality is $K\times M$
to represent an $M$-selection
for the data vector $\mathbf{x}_n$.
$\mathbf{b}_n$ is a concatenation
of $M$ subvectors,
$\mathbf{b}_n = \left[\mathbf{b}_{n1}^T~\mathbf{b}_{n2}^T\cdots\mathbf{b}_{nM}^T\right]^T$.
In each subvector
only one entry $\mathbf{b}_{nm}$ is valued by $1$
and all others are $0$.
The objective function can be formulated as follows,
\begin{align}
\min_{\mathbf{C}, \mathbf{b}_1, \cdots, \mathbf{b}_N}~& ~\sum\nolimits_{n=1}^N \|\mathbf{x}_n - \left[\mathbf{C} \mathbf{C} \cdots \mathbf{C}\right]\mathbf{b}_n\|_2^2. \label{eqn:homogeneouscompositecoding}
\end{align}



Furthermore,
we extend the $M$-selection scheme
to a so-called group $M$-selection scheme.
Group $M$-selection
is a combination of the elements from $M$ sets
$\{\mathbf{C}_1, \cdots, \mathbf{C}_M\}$ each of which is a matrix of size $d \times K$,
taken $1$ at a time from each of the $M$ sets.
The whole formulation is then given as the following,
\begin{align}
\min_{\mathbf{C}_1, \cdots, \mathbf{C}_M, \mathbf{b}_1, \cdots, \mathbf{b}_N}~& ~\sum\nolimits_{n=1}^N \|\mathbf{x}_n - \left[\mathbf{C}_1 \mathbf{C}_2 \cdots \mathbf{C}_M\right]\mathbf{b}_n\|_2^2 \nonumber  \\
\operatorname{s.t.}~& ~ \mathbf{b}_n = \left[\mathbf{b}_{n1}^T~\mathbf{b}_{n2}^T\cdots\mathbf{b}_{nM}^T\right]^T \nonumber  \\
&~ \mathbf{b}_{nm} \in \{0, 1\}^K \nonumber \\
&~ \|\mathbf{b}_{nm}\|_1=1. \label{eqn:complementarycompositecoding}
\end{align}

\begin{figure}[t]
\centering
\subfigure[(a)]{\includegraphics[height=.18\textwidth, clip]{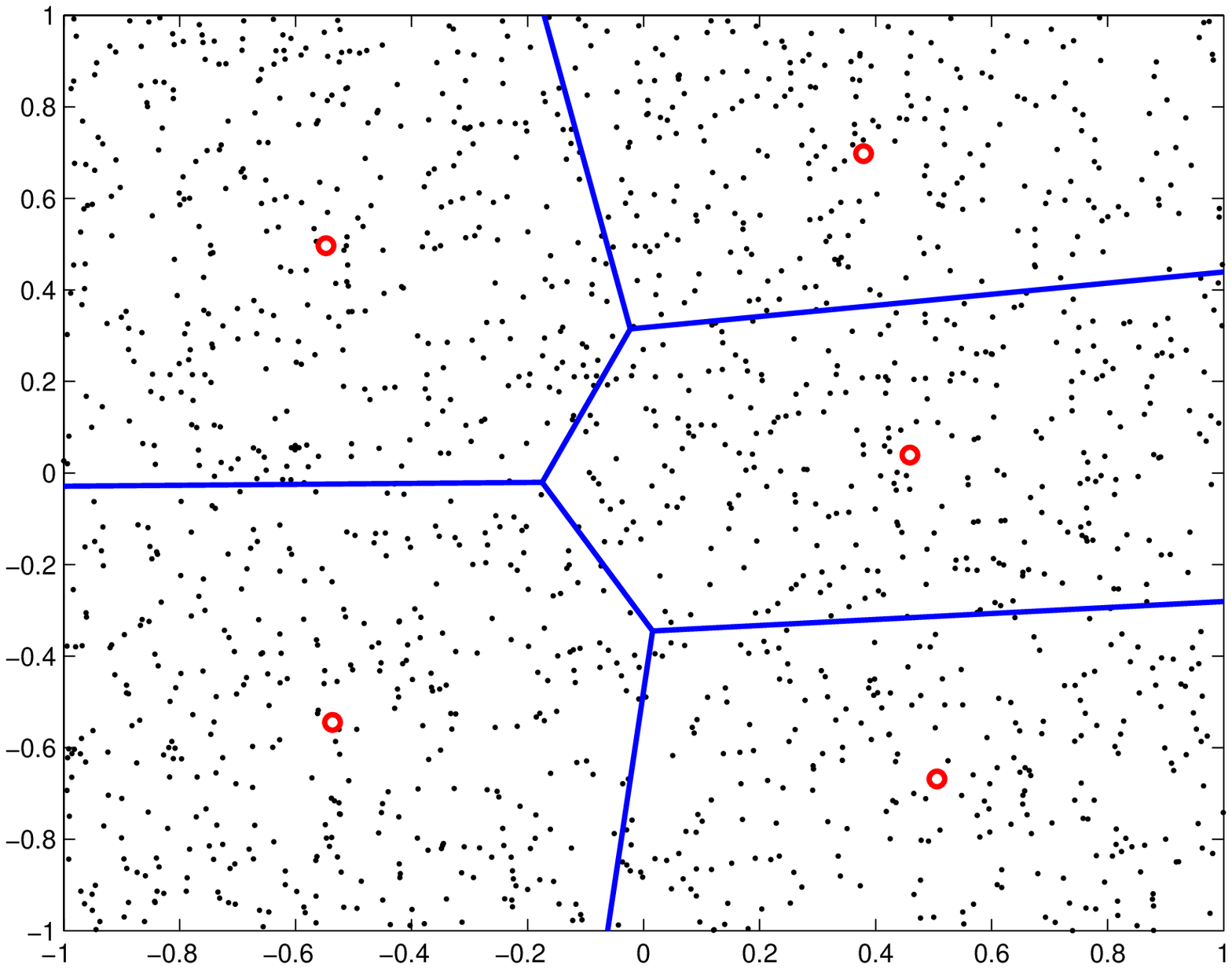}}
\subfigure[(b)]{\includegraphics[height=.18\textwidth, clip]{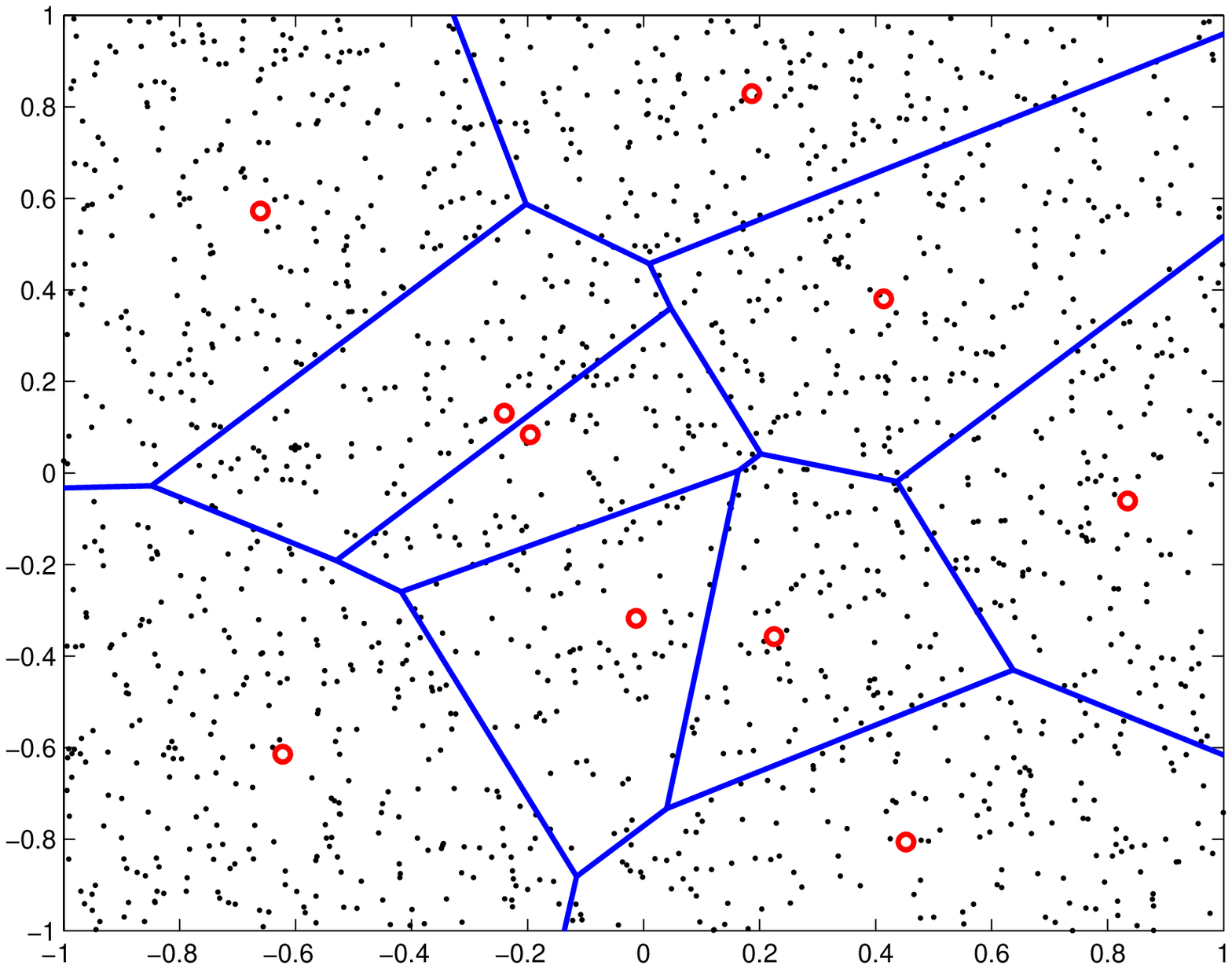}}
\subfigure[(c)]{\includegraphics[height=.18\textwidth, clip]{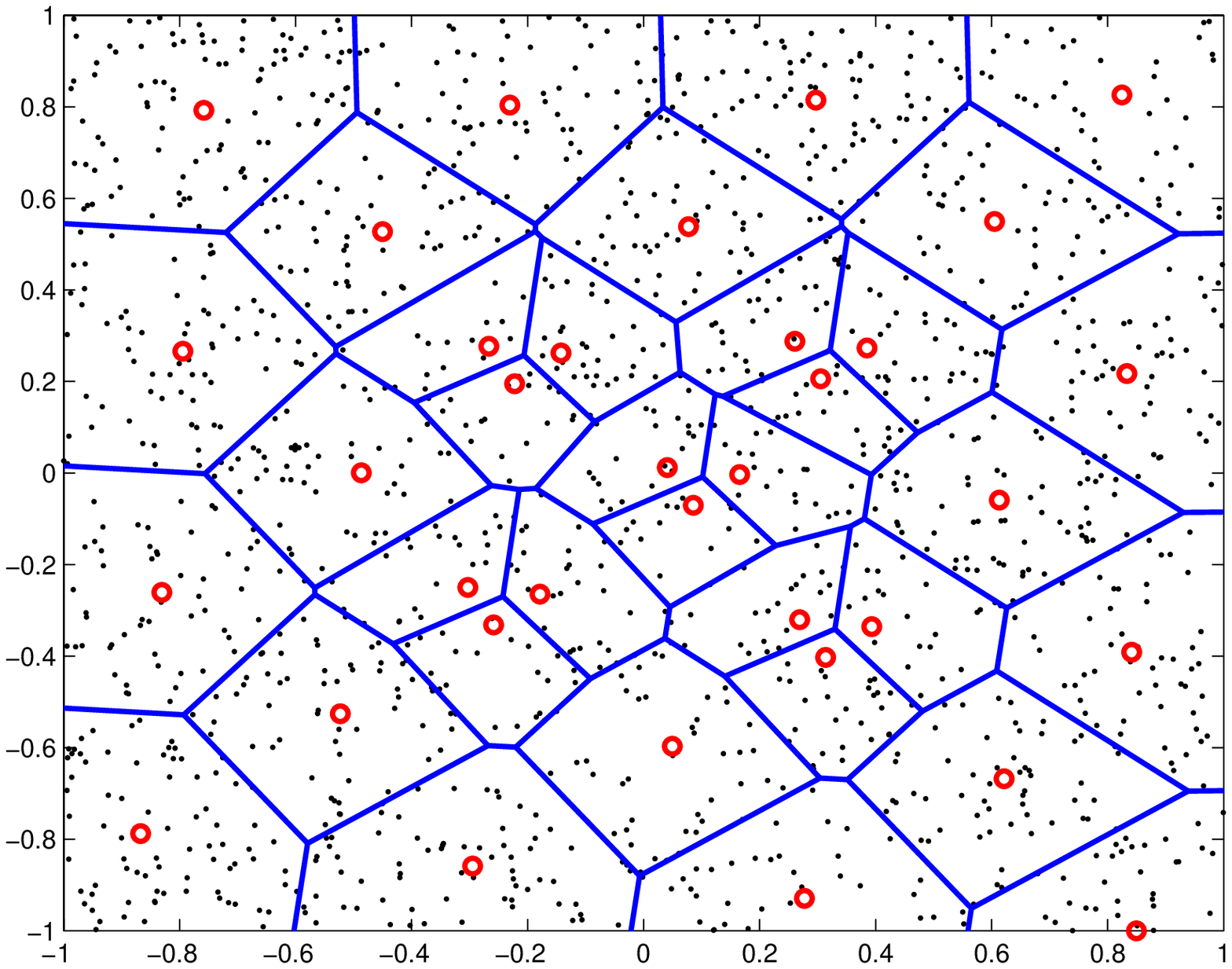}}
\subfigure[(d)]{\includegraphics[height=.18\textwidth, clip]{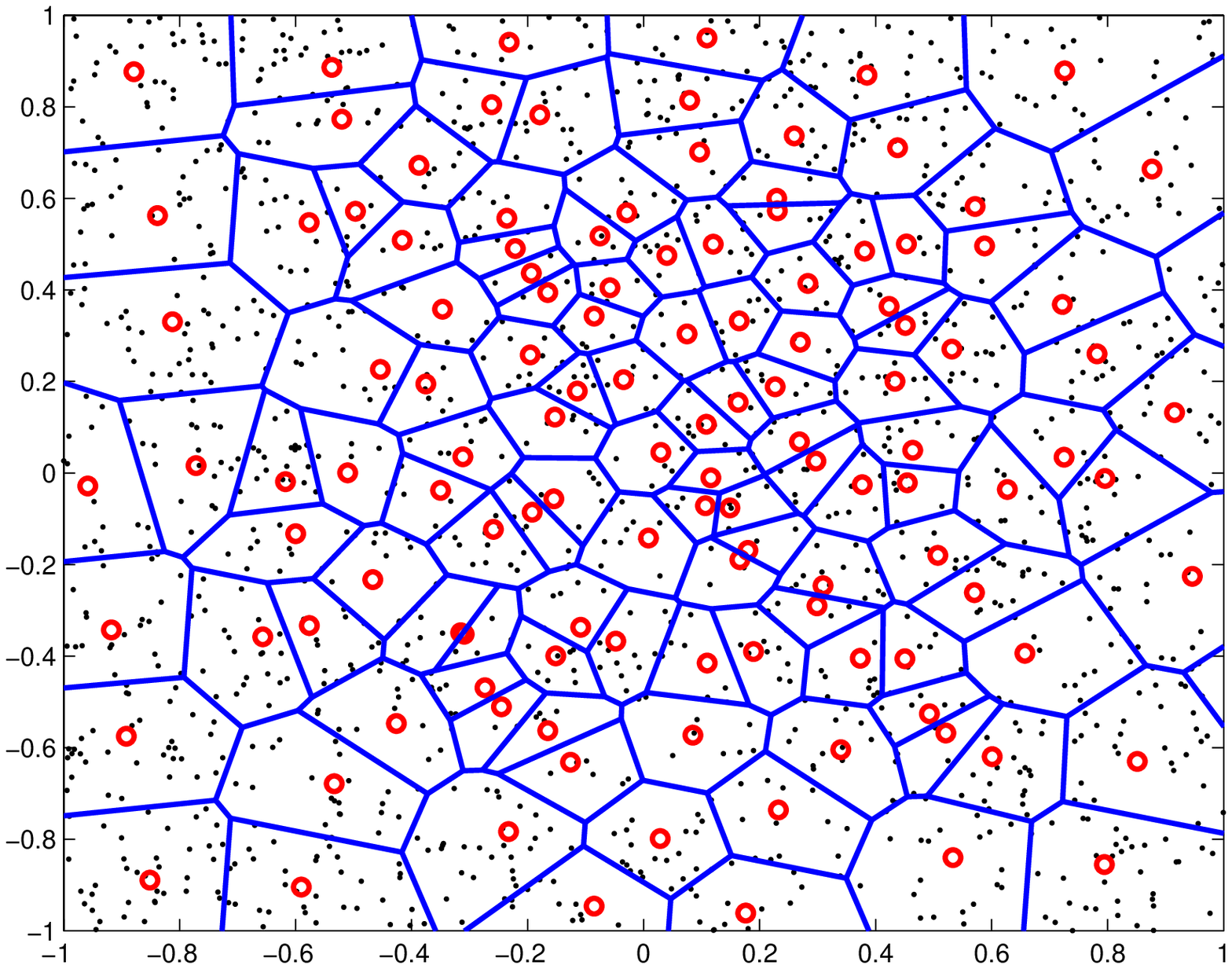}}
\vspace{-0.2cm}
\caption{The illustrations of the compositional dictionaries
learnt with (a) $k$-means,
(b) $M$-combination,
(c) $M$-selection,
and (d) group $M$-selection,
from a set of $1500$ random $2D$ points.
$K=5$, $M=3$.
Each partition corresponds to a dictionary element}
\label{fig:spatialpartitiondifferentselectionschemes}\vspace{-0.5cm}
\end{figure}

The group $M$-selection case,
similar to the $M$-selection case,
also produces an approximation using the composition of $M$ elements,
but differently $M$ source dictionaries
with each containing $K$ elements
are used to form the compositional dictionary.
It, however, does not increase the code length
and keeps the same length $M \log K$,
where each $\mathbf{b}_{nm}$ is represented
by a code of length $\log K$,
denoted by $y_{nm}$
which is
the index of the non-zero entry in $\mathbf{b}_{nm}$
in our implementation.
Note that in this case,
the time consumption for computing the similarity table becomes to $O(MKd)$.
However,
in the large scale similarity search case,
the increase of such computation cost
brought due to the increase of
the number of elements in the source dictionaries
(from $K$ to $KM$)
is neglectable
because $M$ in practice
is chosen to be small
so that the approximate inner product similarity evaluation cost $O(M)$ is not large.
Figure~\ref{fig:spatialpartitiondifferentselectionschemes}
illustrates the compositional dictionaries
from $K$-means, $M$-combination,
$M$-selection and group $M$-selection.

\subsection{Optimization}
We adopt alternating optimization,
which is used in the Lloyd and sparse dictionary learning algorithms,
to solve the optimization problem in Equation~\ref{eqn:complementarycompositecoding}.
To be clear, we denote $\mathbf{D} = [\mathbf{C}_1\cdots\mathbf{C}_M]$.
Our optimization algorithm alternatively optimizes
the source dictionaries $\mathbf{D}$
and optimizes the group $M$-selection $\mathbf{b}_n$ for each data vector $\mathbf{x}_n$.

\noindent\textbf{Update the dictionary.}
The objective function can be transformed
as $f(\mathbf{D}, \mathbf{B}) = \|\mathbf{X} - \mathbf{D}\mathbf{B}\|_F^2$,
where $\mathbf{X} = [\mathbf{x}_1\cdots \mathbf{x}_N]$
is the data matrix
and $\mathbf{B} = [\mathbf{b}_1 \cdots \mathbf{b}_N]$
is the group $M$-selection matrix.
This is a quadratic optimization problem
w.r.t. the variable $\mathbf{D}$.
Many algorithms have been designed to solve this problem.
In this paper,
we solve this problem using the closed-form solution.
Let the derivative of $f(\mathbf{D}, \mathbf{B})$
with respect to $\mathbf{D}$ be zero:
$\frac{\partial f(\mathbf{D}, \mathbf{B})}{ \partial \mathbf{D}}
= 2 (\mathbf{D}\mathbf{B}\mathbf{B}^T - \mathbf{X}\mathbf{B}^T) = 0$.
Then we have the closed-form solution:
$\mathbf{D} = \mathbf{X}\mathbf{B}^T(\mathbf{B}\mathbf{B}^T)^{-1}$.
The online learning algorithm~\cite{MairalBPS09}
can be also be borrowed for acceleration.

\noindent\textbf{Update the group $M$-selection $\mathbf{b}_n$.}
Optimizing $\{\mathbf{b}_1, \cdots, \mathbf{b}_N\}$
given the source dictionaries $\mathbf{D}$
can be decomposed into $N$ independent subproblems
$\{\min_{\mathbf{b}_n} (f_n(\mathbf{b}_n) =  \|\mathbf{x}_n - \mathbf{D}\mathbf{b}_n\|_2^2)\}_{n=1}^N$
with the associated constraints,
each of which optimizes the group $M$-selection $\mathbf{b}_n$ separately.

Intuitively, each subproblem selects one element from each
source dictionary so that the composition of these elements is
the closest to the data vector. The subproblem of minimizing
$f_n(\mathbf{b}_n)$ is a combinatorial problem and generally
NP-hard. We propose to adopt a greedy algorithm, performing $M$
$1$-selection optimizations over the $M$ source dictionaries in
the best-first manner.
The $m$-th iteration of the greedy
algorithm consists of
determining over which source dictionary
$1$-selection is performed
from the remaining $(M-m-1)$ source
dictionaries that have not been selected and finding the best
$1$-selection over the selected source dictionary. The former
issue is solved by selecting the best source dictionary over
which the reconstruction error given the previous selected
$(m-1)$ $1$-selections is the minimal. The latter issue is
solved by selecting the element in the source dictionary that
is the nearest to the residual (the difference of the data
vector from the current approximate vector using the previous
$(m-1)$ $1$-selections).

\subsection{Search with Compositional Codes}\label{subsection:searchwithcc}
Given that the database $\{\mathbf{x}_n\}_{n=1}^N$
is represented by the compact codes
$\{\mathbf{y}_n\}_{n=1}^N$
with $\mathbf{y}$ being $M$ codes
$y_{n1}, y_{n2}, \cdots, y_{nM}$,
we perform the linear scan search to find nearest neighbors,
by computing the approximate similarity
of a query $\mathbf{q}$
with each database vector.
The inner product similarity between $\mathbf{q}$ and $\mathbf{x}_n$
is approximated using the compact code,
$\texttt{<}\mathbf{q}, \mathbf{x}_n\texttt{>} \approx
\texttt{<}\mathbf{q}, \sum_{m=1}^M \mathbf{c}_{y_{nm}}\texttt{>}$.
The distributive property shows that
$\texttt{<}\mathbf{q}, \sum_{m=1}^M \mathbf{c}_{y_{nm}}\texttt{>}
= \sum_{m=1}^M \texttt{<}\mathbf{q}, \mathbf{c}_{y_{nm}}\texttt{>}$,
which takes $O(M)$ time
if the inner products between the query and the dictionary elements
have been already computed.
As aforementioned, before linear scan,
the search process first constructs the similarity table,
storing inner products between the query $\mathbf{q}$
and all the dictionary elements in $\mathcal{C}$, whose time complexity is $O(MKd)$.
In summary, the overall search time complexity is
$O(MKd + NM)$.
When handling large scale data,
the cost of computing the similarity table is relatively small
and neglectable compared with the linear scan cost.
In the case of searching over SIFT$1M$ with $M=8$ and $K=256$,
the time of computing the similarity table is about only $2\%$
of the total search time
and in the case of searching over SIFT$1B$ with $M=8$ and $K=256$,
the ratio for the cost of similarity table computation is even much smaller.

\subsection{Analysis}
Let's see how to transform the group $M$-selection case
formulated in Equation~\ref{eqn:complementarycompositecoding}
to other cases.
We introduce three constraints:
$c1$: $\mathbf{C}_1=\mathbf{C}_2=\cdots = \mathbf{C}_M$;
$c2$: $\mathbf{b}_{ni} \neq \mathbf{b}_{nj}, \forall i \neq j, \forall n$;
$c3$: $M=1$.
It is easy to show that
the formulation in Equation~\ref{eqn:complementarycompositecoding}
with an extra constraint $c1$
is equivalent to the $M$-selection case,
that it with two extra constraints $c1$ and $c2$ is reduced to
the $M$-combination case,
and that it together with all the three extra constraints,
$c1$, $c2$ and $c3$,
is reduced to the $k$-means case.
The reduction relations are summarized
as the following property.

\begin{Property}
\label{property:relationsofformulations}
The compositional code approach with group $M$-selection
can be transformed
to the ones with $M$-selection
and $M$-combination
and $k$-means
by successively adding extra constraints:
Group $M$-selection
 $\underrightarrow{+c1}$
$M$-selection $\underrightarrow{+c2}$
$M$-combination $\underrightarrow{+c3}$
$k$-means.
\end{Property}

With regard to the optimal objective function values of
$k$-means,
$M$-combination,
$M$-selection,
and group $M$-selection
that are denoted by
$f^*_{km}$,
$f^*_{mc}$,
$f^*_{ms}$,
and
$f^*_{gms}$,
respectively,
we have the following property.

\begin{Property}
\label{property:relationsofobjectivevalues}
Given the same database $\mathcal{X}$
and the same variables of $K$ and $M$,
we have
(1) $f^*_{gms} \leqslant f^*_{ms}$;
(2) $f^*_{ms} \leqslant  f^*_{mc}$;
(3) $f^*_{ms} \leqslant  f^*_{km}$.
There is no guarantee
for $f^*_{mc} \leqslant  f^*_{km}$.
\end{Property}

The proofs of the first three inequalities
in the above property is obvious
and it is easily validated
that the optimal solution of the $M$-selection ($M$-combination, $k$-means) case
is (or forms) a feasible solution of the group $M$-selection ($M$-selection, $M$-selection) case.
In contrast,
we can find an example
that the optimal solution of the $k$-means case
cannot form a feasible solution of the $M$-combination case (e.g., in the case $K=M$).

We compute
the cardinalities of the compositional dictionaries
(the source dictionary in $k$-means
is equivalently regarded as the compositional dictionary)
in the four cases
to show the difference of the four algorithms in another way.
Generally,
the objective value would be smaller
if the cardinality of the compositional dictionary is larger.
The cardinalities are summarized as follows.


\newcommand{\multibinom}[2]{
  \big(\!\big(\genfrac{}{}{0pt}{}{#1}{#2}\big)\!\big)
}

\begin{Property}
\label{property:cardinalities}
The cardinalates of the group $M$-selection case,
the $M$-selection case,
the $M$-combination case,
and the $k$-means case
are $K^M$,
$\multibinom{K}{M} = \binom{K+M-1}{M}
=\frac{(K+M-1)!}{M!(K-1)!}$,
$\binom{K}{M} = \frac{K!}{M!(K-M)!}$,
and $K$,
respectively.
We have
$K^M \geqslant \multibinom{K}{M} \geqslant
\binom{K}{M}$,
and
$K^M \geqslant \multibinom{K}{M} \geqslant
K$.
\end{Property}

Property~\ref{property:relationsofobjectivevalues} shows
that the group $M$-selection scheme produces
the smallest objective values.
In other words,
the group $M$-selection scheme leads to the most accurate approximation
on average.
Based on the bound analysis
in Property~\ref{property:innerproductbound},
it can be concluded
that group $M$-selection
can achieve the most accurate inner product approximation
as given in the following corollary.

\begin{Corollary}
\label{corollary:gruopmselectionisbest}
On average, the group $M$-selection scheme
results in more accurate inner product approximation
(smaller error upper-bound)
than $k$-means, $M$-combination, and $M$-selection
given the same variables $M$ and $K$.
\end{Corollary}

\noindent\textbf{Time Complexity.}
Section~\ref{subsection:searchwithcc} has described the search process
and its time complexity.
The following
presents the time complexity for the training process.
The training process is an iterative procedure
and each iteration consists of two steps:
dictionary learning
and code updating.
The dictionary learning step updates the dictionaries
as a closed-form solution:
$\mathbf{D} = \mathbf{X}\mathbf{B}^T(\mathbf{B}\mathbf{B}^T)^{-1}$,
which includes (sparse) matrix multiplication and matrix inversion,
and its time complexity is
$O(NMd + d (MK)^2 + NM^2 + (MK)^3)$.
The code updating step involves computing the code for each vector, taking $O(M^2Kd)$,
and thus takes $O(NM^2Kd)$ for all database vectors.
In a word,
the time complexity of
the whole iteration process
is $O(T(NM^2Kd + d (MK)^2 + (MK)^3))$
with $T$ being the number of iterations
and one can see that it is linear with respect to the number of vectors $N$
and the dimension $d$.
when training the codes for the SIFT$1M$ dataset,
the algorithm reaches convergence in $15$ iterations
and takes about $250 \times 15$ seconds (with a single Intel i$7$-$2600$ CPU ($3.40G$Hz))
Our algorithm also benefits from parallel computing,
and thus the practical time consumption is acceptable,
for example,
computing the codes from the $100M$ learning vectors for SIFT$1B$
is completed within $5$ hours.

\noindent\textbf{Connections and Discussions.}
We summarize the relations with several closely-related algorithms.
Detailed analysis
is given in the supplementary material.
Product quantization~\cite{JegouDS11} and Cartesian $k$-means~\cite{NorouziF13}
can be viewed as a constrained version of the group $M$-selection algorithm:
each subquantizer corresponds to
a source dictionary in our approach
and each source dictionary lies in a different subspace
with the same dimension (in the case that each subspace is full-ranked).
In comparison with order permutation~\cite{ChavezFN08}
in which the similarity between permutation orders
is used as a predictor of the closeness,
our approach uses the composition of selected dictionary elements
to approximate the vector
and thus uses it for inner product similarity approximation.

The proposed $M$-selection and group $M$-selection schemes
can be regarded as a sparse coding approach
with group sparsity~\cite{YuanL06} in which the coefficients
are divided into several groups
and the sparsity constraints are imposed in each group separately.
In particular,
the coefficients in our approach that can be only
valued by $0$ or $1$
are divided into $M$ groups
and for each group the non-sparsity degree is $1$.

\section{Experiments}
\setlength{\tabcolsep}{4pt}
\begin{table}[t]
\begin{center}
\caption{The descriptions of the four datasets. \#(database) =
\#(database vectors). \#(queries) = \#(query vectors)}
\label{tab:dataset}
\footnotesize
\begin{tabular}{l|llll}
\hline
& SIFT1M & SIFT1B & Netflix & LM$1M$ \\
\hline
dimension & $128$ & $128$ & $17,770$ & $2,048$ \\
\#(database) & $1,000,000$ & $1,000,000,000$ & $480,189$ & $890,912$ \\
\#(queries) & $10,000$ & $10,000$ & $480,189$ & $100,000$ \\
\hline
\end{tabular}
\vspace{-.5cm}
\end{center}
\end{table}
\setlength{\tabcolsep}{1.4pt}

We conduct the inner product similarity search experiments
over four data sets:
SIFT$1M$~\cite{JegouDS11},
SIFT$1B$~\cite{JegouTDA11},
linear models (LM$1M$),
and Netflix.
The SIFT$1M$ dataset
consists of $1M$ $128$-dimensional SIFT descriptors
as the database vectors
and $10K$ SIFT descriptors
as the query vectors,
which
are extracted from the INRIA
holidays images~\cite{JegouDS08}.
The SIFT$1B$ dataset contains $1$ billion
SIFT features as the database vectors,
$100M$ SIFT features as the learn vectors
and $10K$ SIFT features
as the query vectors,
which
are extracted from approximately $1$ million images.
The LM$1M$ dataset consists of
around $1M$ ($890, 912$) linear models
with the weight vector as the database vectors,
which are
learnt from $890, 912$ textual queries
with the images frequently clicked in a commercial search engine
for each textual query
as the training samples
using the PAMIR approach~\cite{GrangierB08}
and $100K$ $2048$-dimensional image features
as the queries.
Netflix~\cite{BennettL07} contains
a rating matrix that $480,189$ users gave to $17,770$ movies
and aims to predict user ratings for films.
It is shown that
inner product between the rating vectors that two users
gave to $17,770$ movies
can be used to evaluate
the similarity of users' interest
which can help recommend films to users.
In our experiments,
the dimension of the rating vector is reduced to $512$
with PCA.
The descriptions of the datasets
are summarized in Table~\ref{tab:dataset}.

The search quality is measured with recall$@P$,
which is defined as
the fraction of relevant instances that are retrieved
among the first $P$ positions.
Relevant instances in our case
are $R$-nearest neighbors
under the inner product similarity.
This measure is equivalent to
the precision measure
if $R=P$,
or when evaluating the returned top $R$ results
after performing a subsequent reranking scheme using the exact inner product similarity
over the retrieved $P$ items,
following an approximate search step.
The true nearest neighbors under the inner product similarity
are computed by comparing each query with all the database vectors
using the raw features.

\subsection{Empirical Analysis}
\begin{figure}[t]
\centering
\includegraphics[height=.26\textwidth, clip]{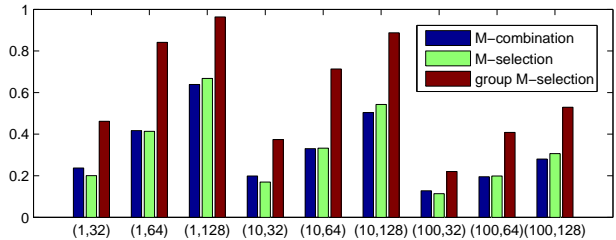}
\vspace{-0.2cm}
\caption{Performance comparison
of $M$-combination, $M$-selection,
and group $M$-selection.
$(x, y) = (\text{\#(NNs)}, \text{\#(bits)})$}
\label{fig:comparisondifferentselectionschemes}\vspace{-0.5cm}
\end{figure}

We report the performance of compositional code
using the proposed three schemes:
$M$-combination,
$M$-selection,
and group $M$-selection.
The results of searching different numbers of nearest neighbors
with different numbers of bits
at the same position $100$
are shown in Figure~\ref{fig:comparisondifferentselectionschemes}.
This result and all the following results
are obtained by fixing $K=256$
(i.e., each source dictionary is encoded with a byte)
and tuning $M$ to vary the number of bits.
We can see that group $M$-selection
performs the best,
which is consistent to the analysis
that group $M$-selection is the best
on average in vector approximation
and inner product approximation.
In addition,
from the results in Figure~\ref{fig:comparisondifferentselectionschemes},
it can be observed that
more bits result in better performance
when searching the same number of nearest neighbors
and the performance when searching more nearest neighbors
with the same number of bits
decreases.

\begin{figure*}[t]
\centering
\subfigure{\includegraphics[width=.29\textwidth, clip]{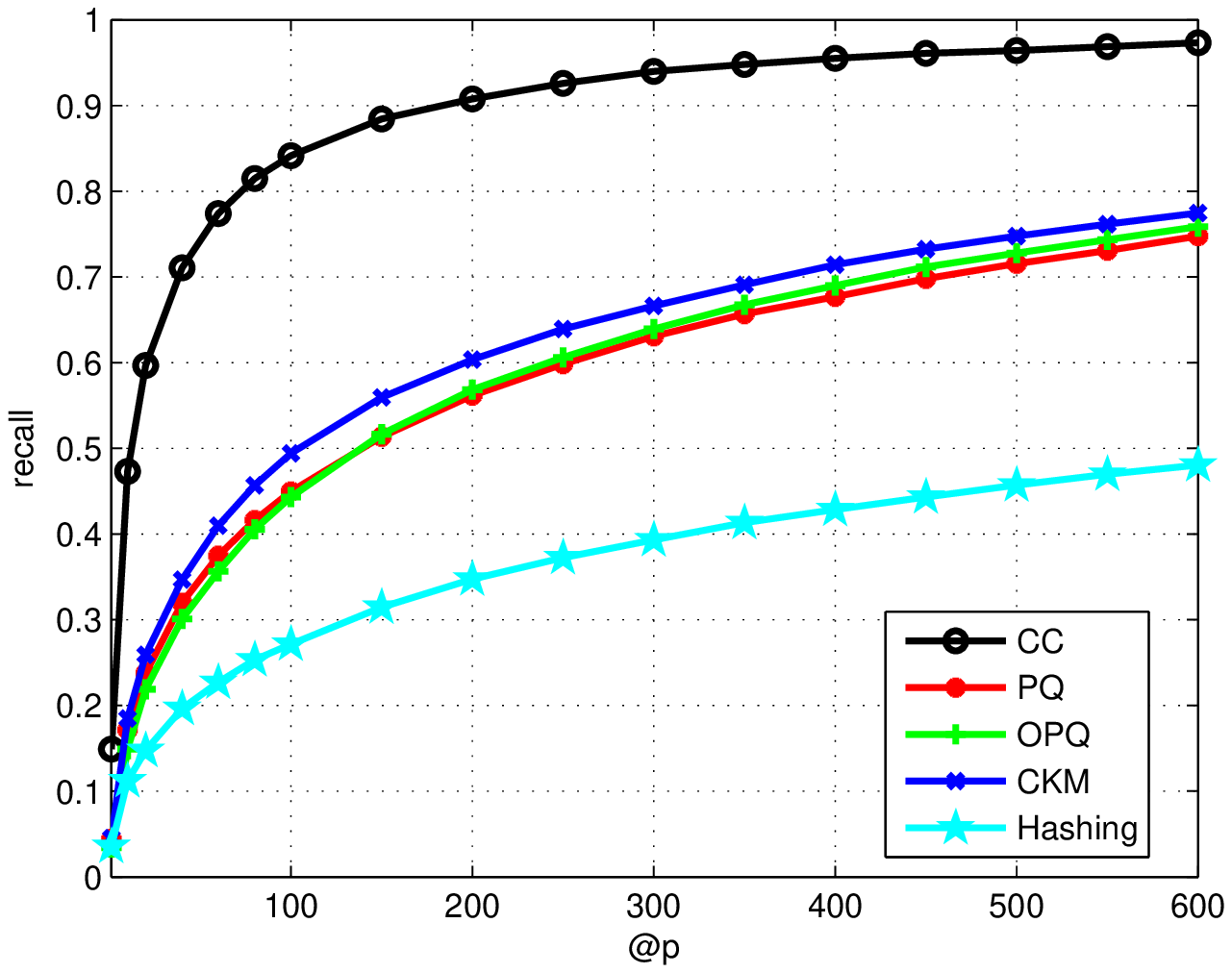}}~~~~~
\subfigure{\includegraphics[width=.29\textwidth, clip]{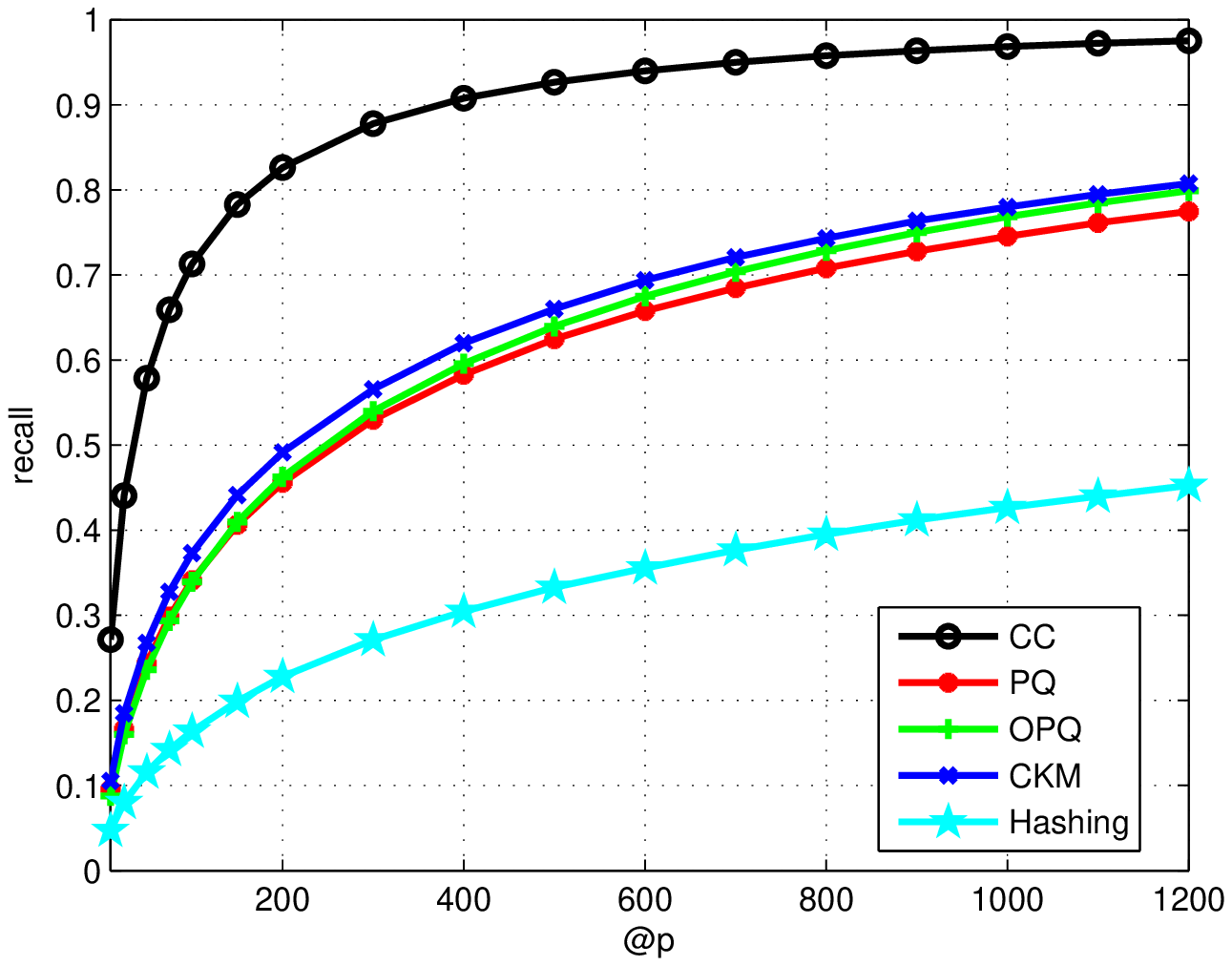}}~~~~~
\subfigure{\includegraphics[width=.29\textwidth, clip]{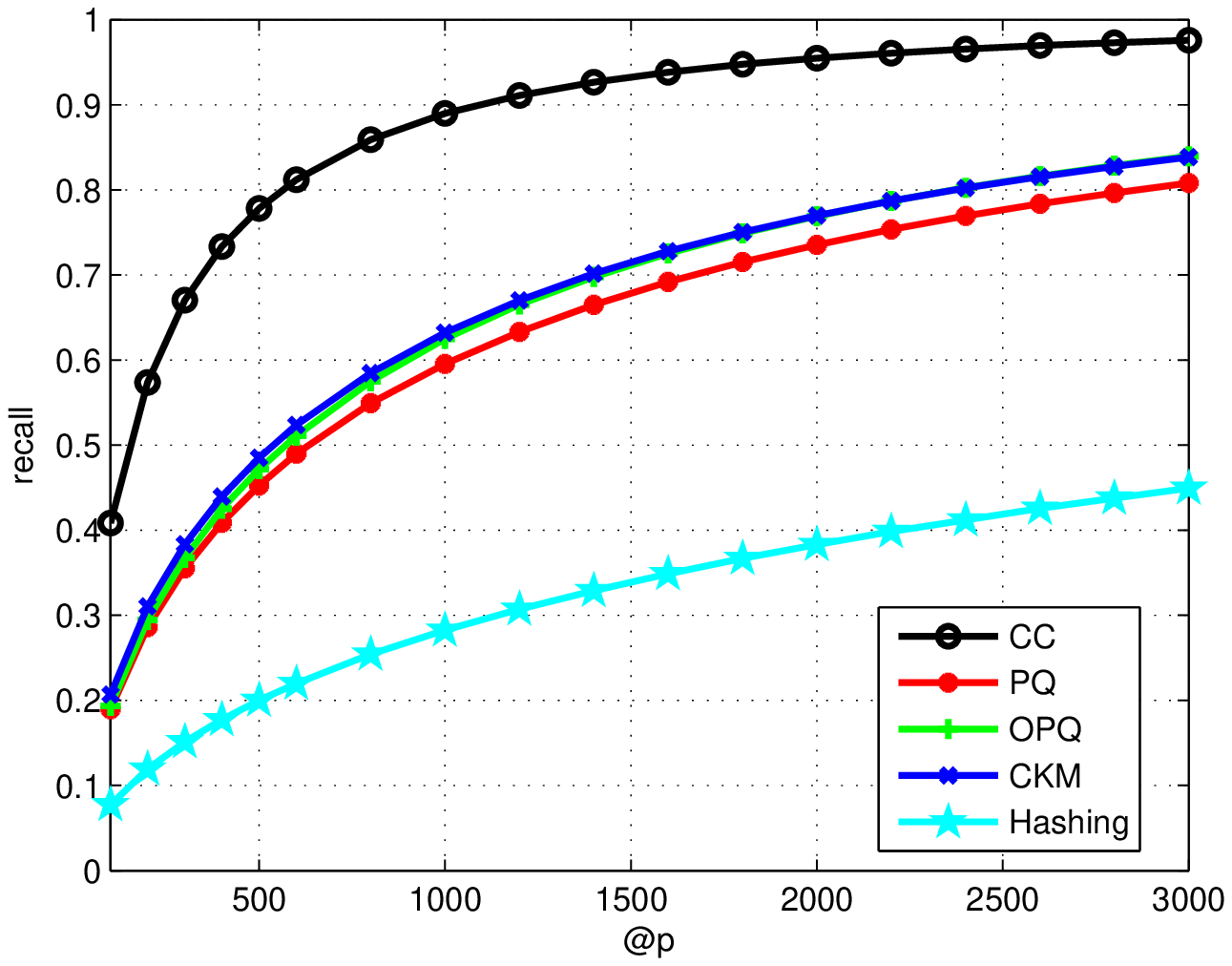}}\\
\vspace{-.2cm}
\subfigure[(a)]{\includegraphics[width=.29\textwidth, clip]{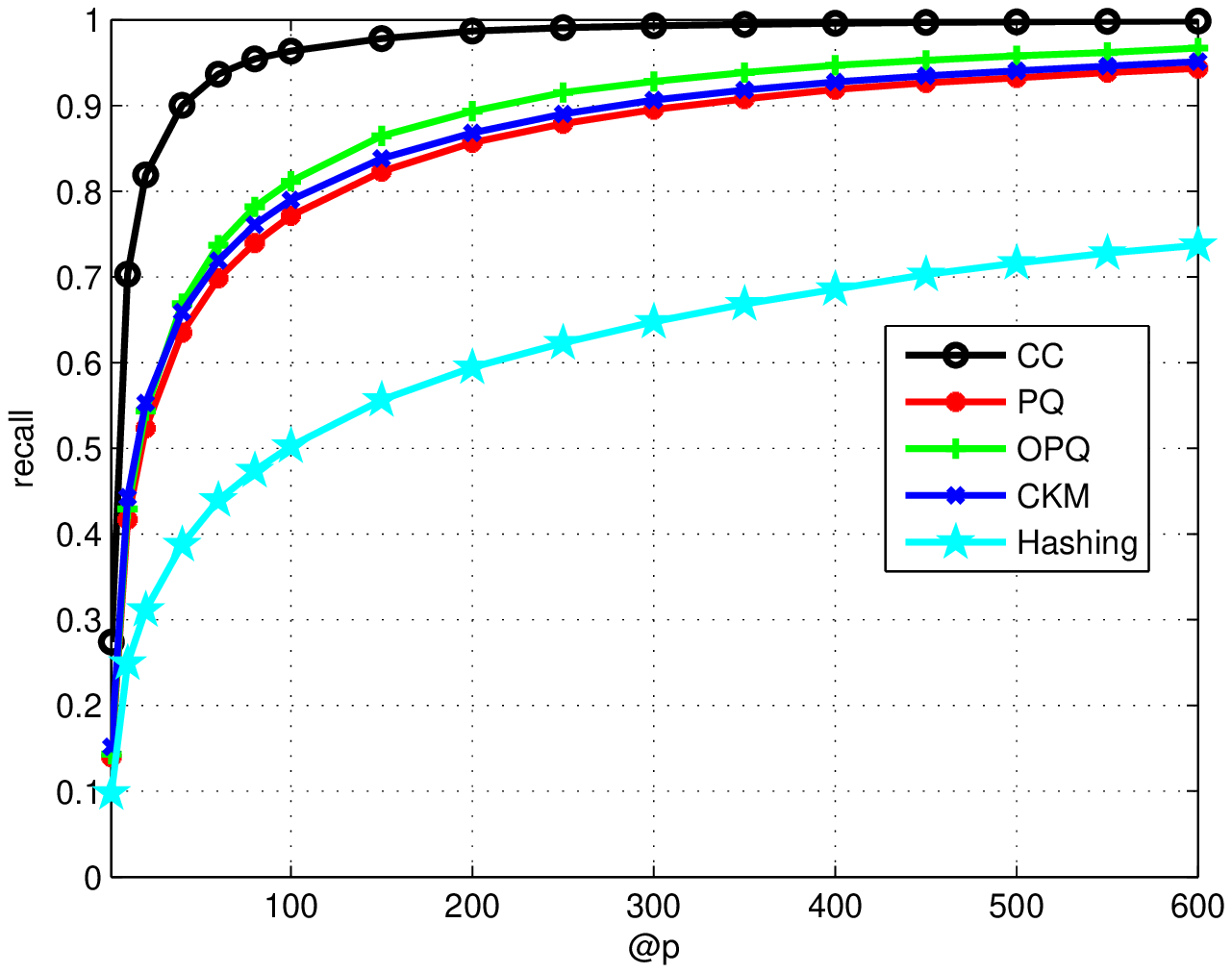}}~~~~~
\subfigure[(b)]{\includegraphics[width=.29\textwidth, clip]{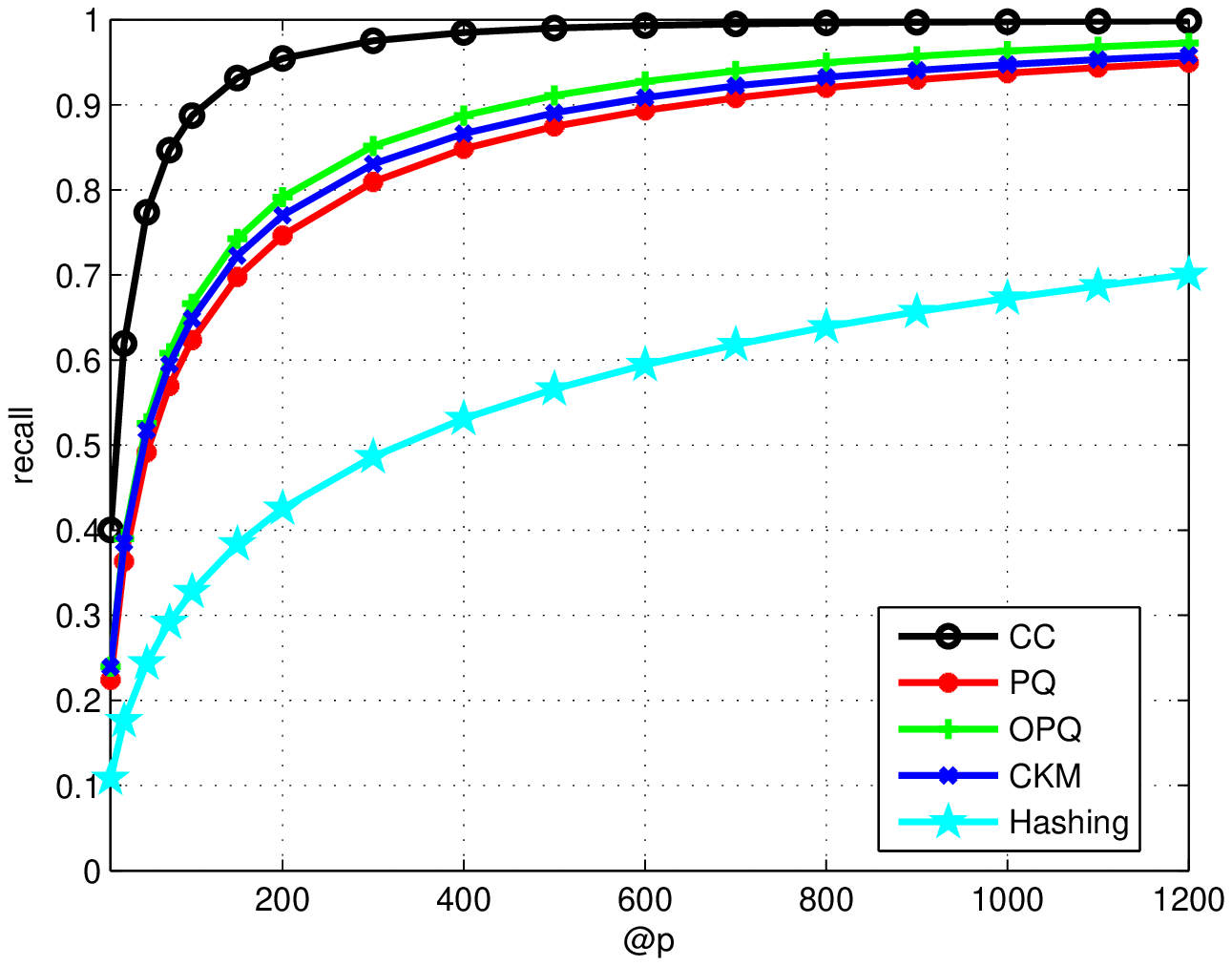}}~~~~~
\subfigure[(c)]{\includegraphics[width=.29\textwidth, clip]{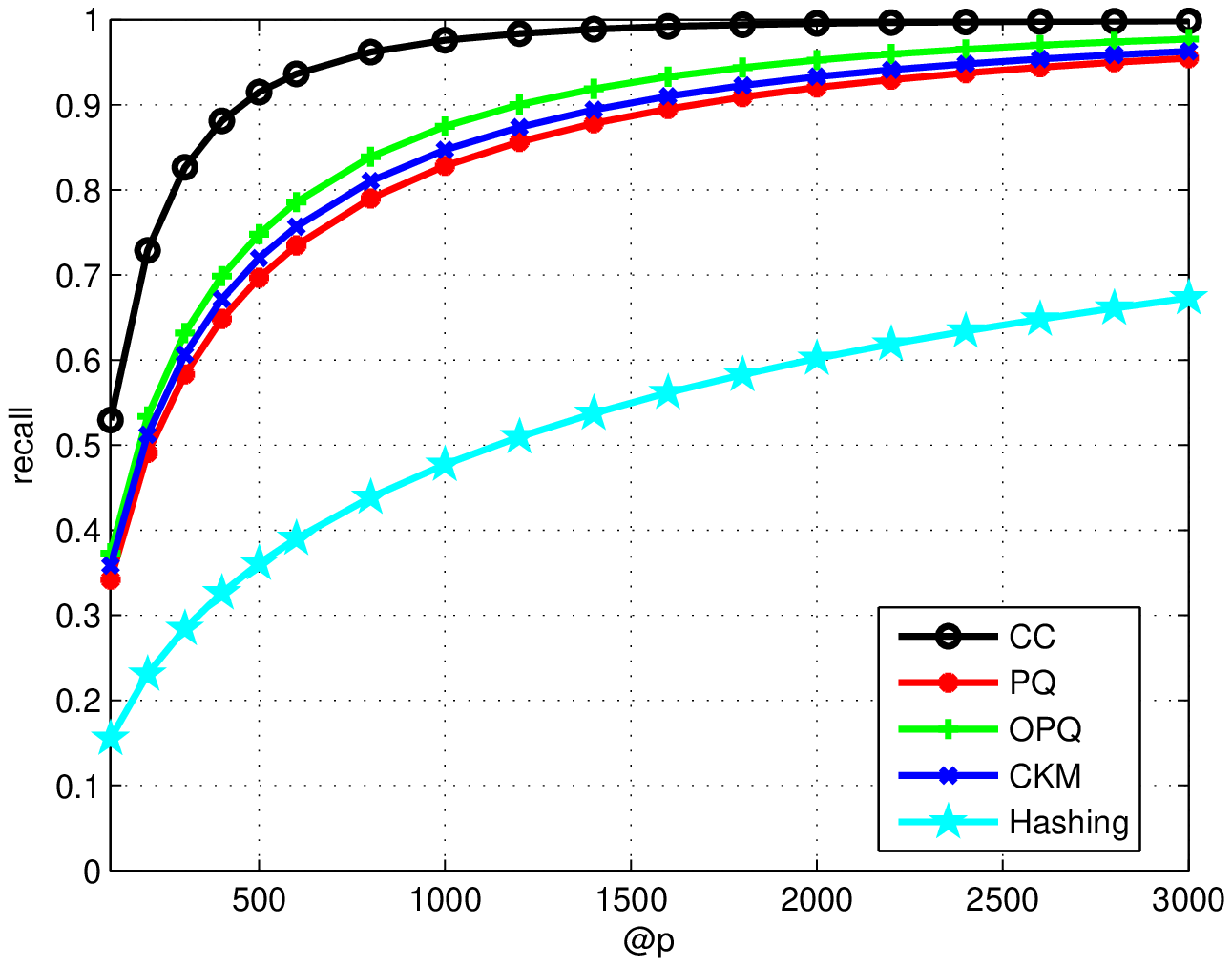}}
\vspace{-0.2cm}
\caption{Performance comparison over SIFT$1M$ with $64$ (top) and $128$ (bottom) bits
for searching (a) $1$-NN, (b) $10$-NNs, and (c) $100$-NNs}
\label{fig:SIFT1MResultsComparison}\vspace{-0.5cm}
\end{figure*}

\begin{figure*}[t]
\centering
\subfigure{\includegraphics[width=.29\textwidth, clip]{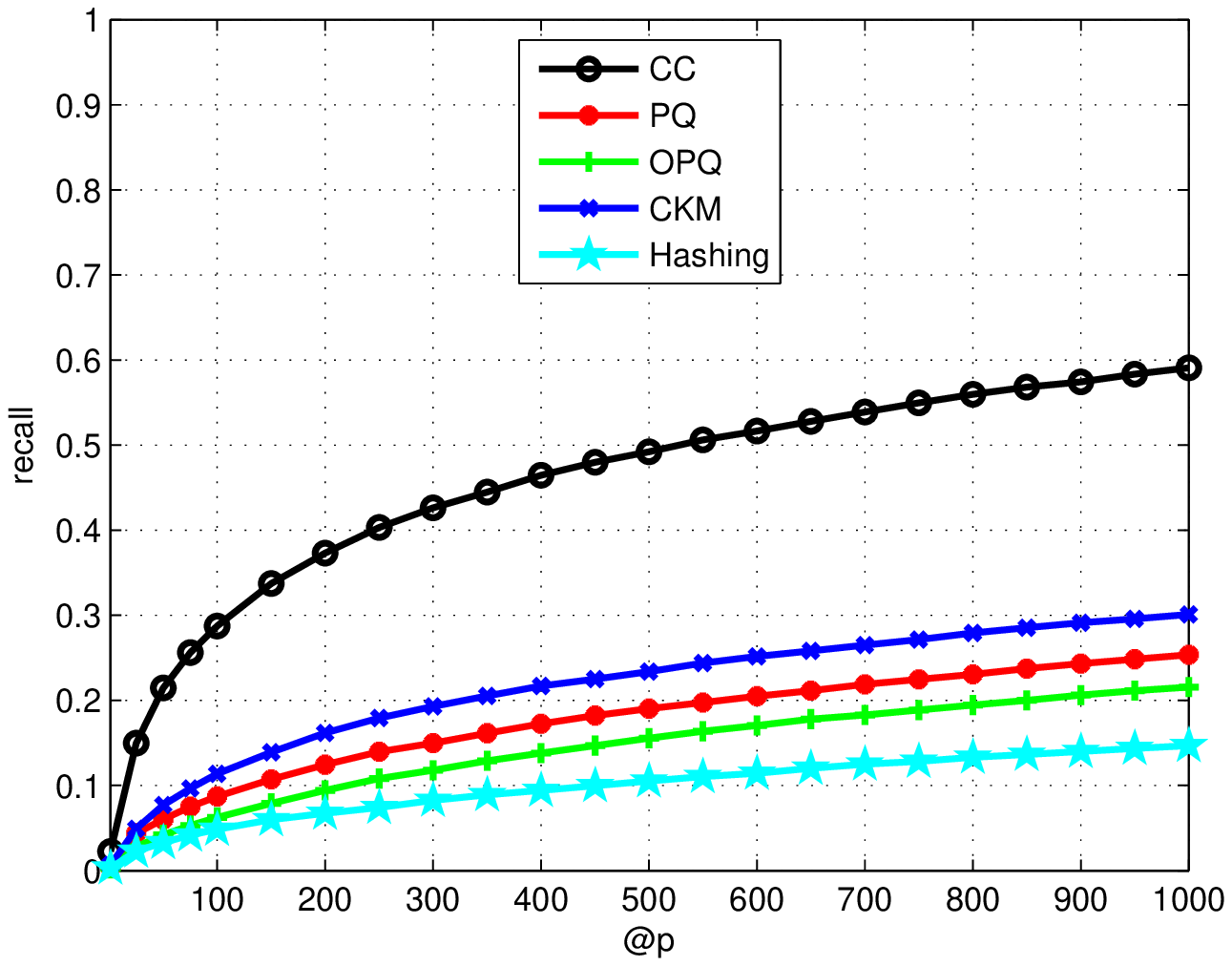}}~~~~~
\subfigure{\label{fig:subfig:b}\includegraphics[width=.29\textwidth, clip]{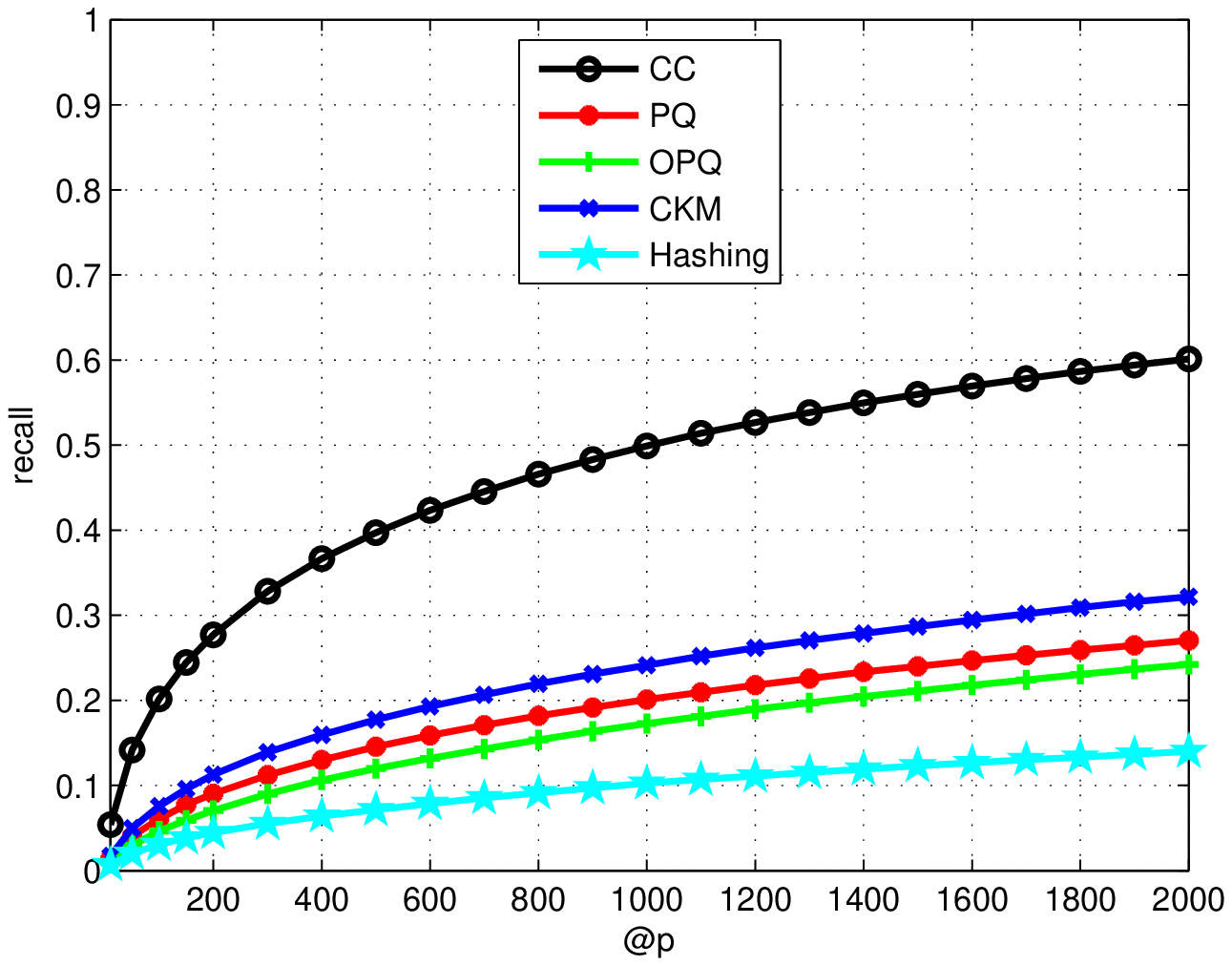}}~~~~~
\subfigure{\includegraphics[width=.29\textwidth, clip]{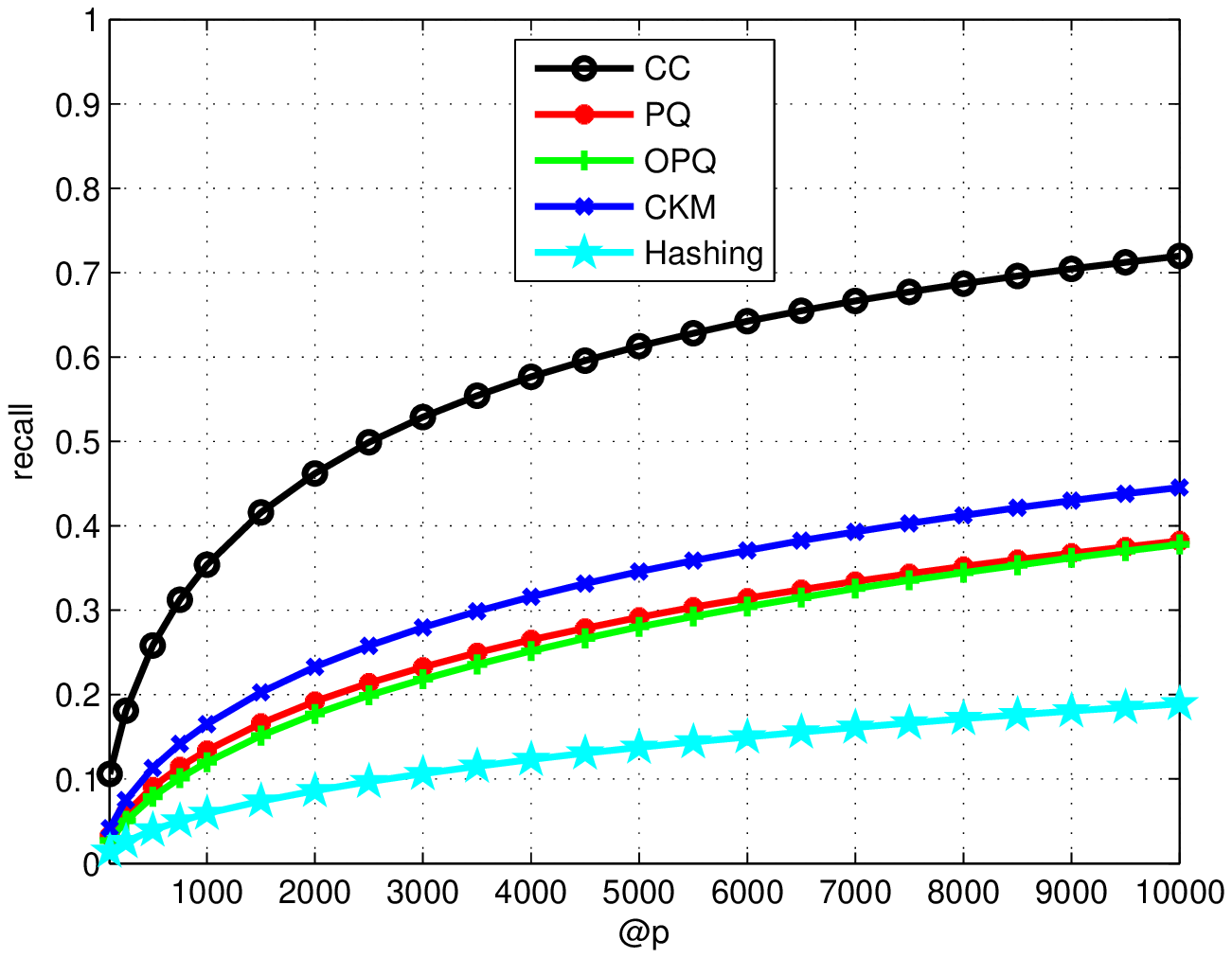}}\\
\vspace{-.2cm}
\subfigure[(a)]{\includegraphics[width=.29\textwidth, clip]{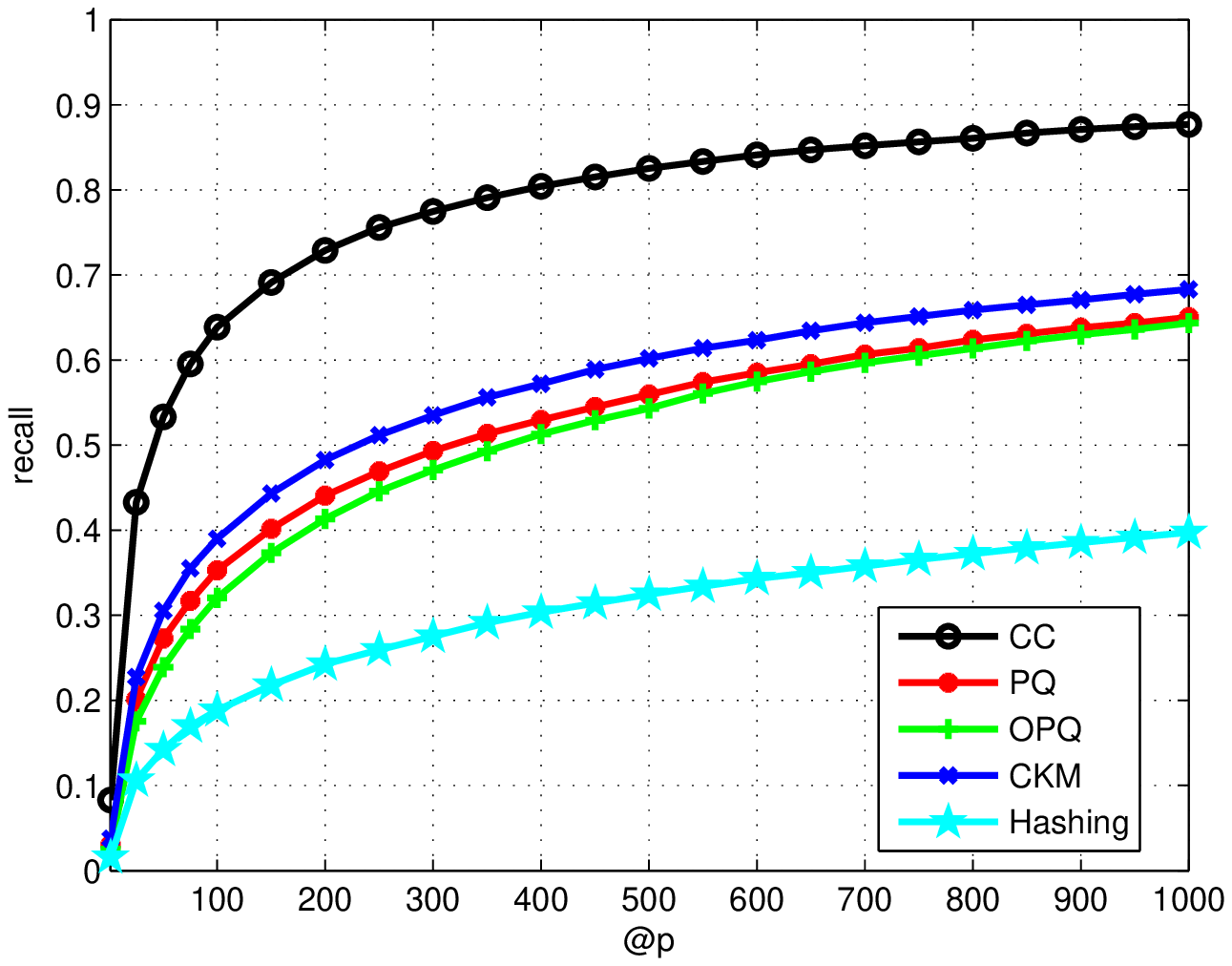}}~~~~~
\subfigure[(b)]{\includegraphics[width=.29\textwidth, clip]{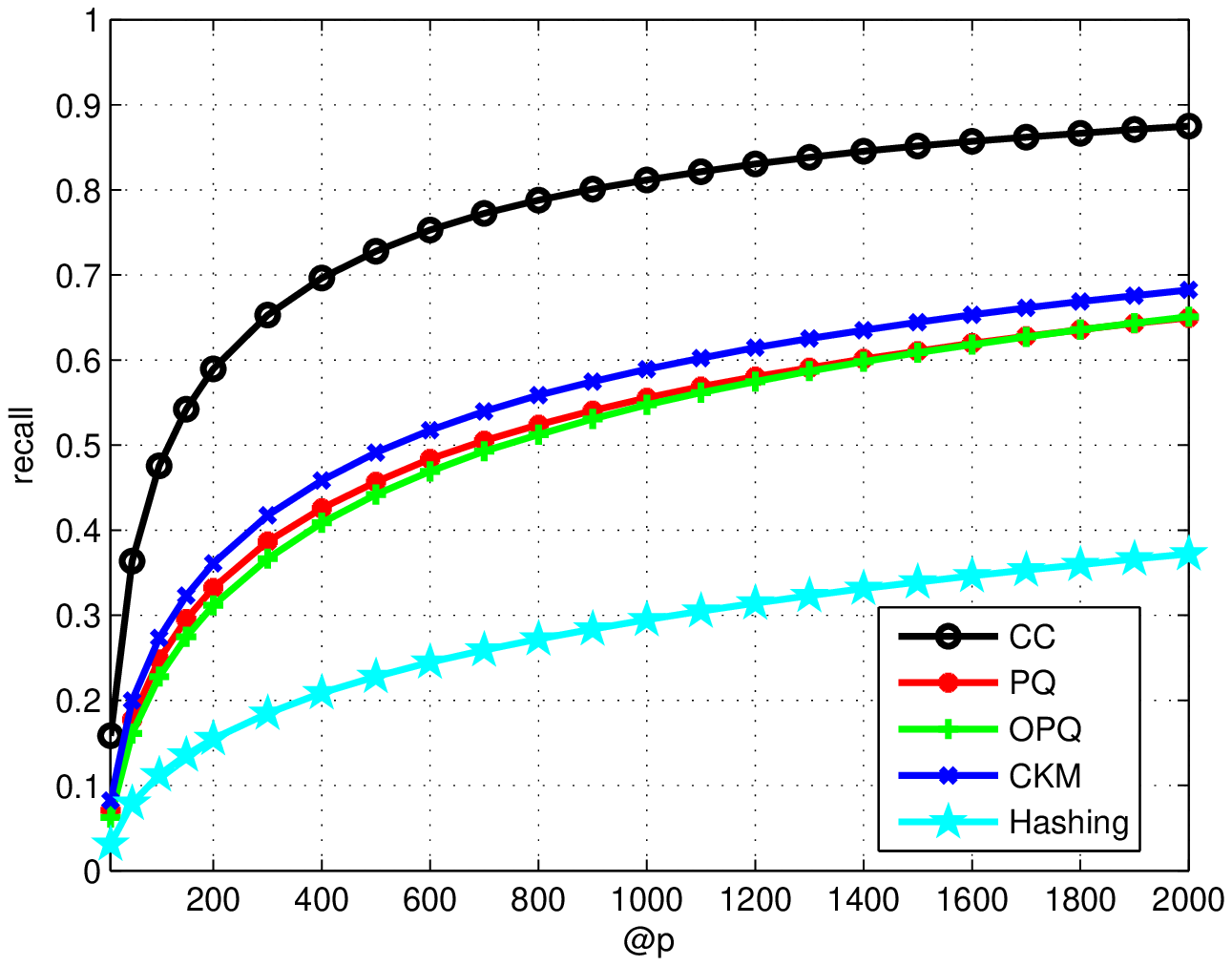}}~~~~~
\subfigure[(c)]{\includegraphics[width=.29\textwidth, clip]{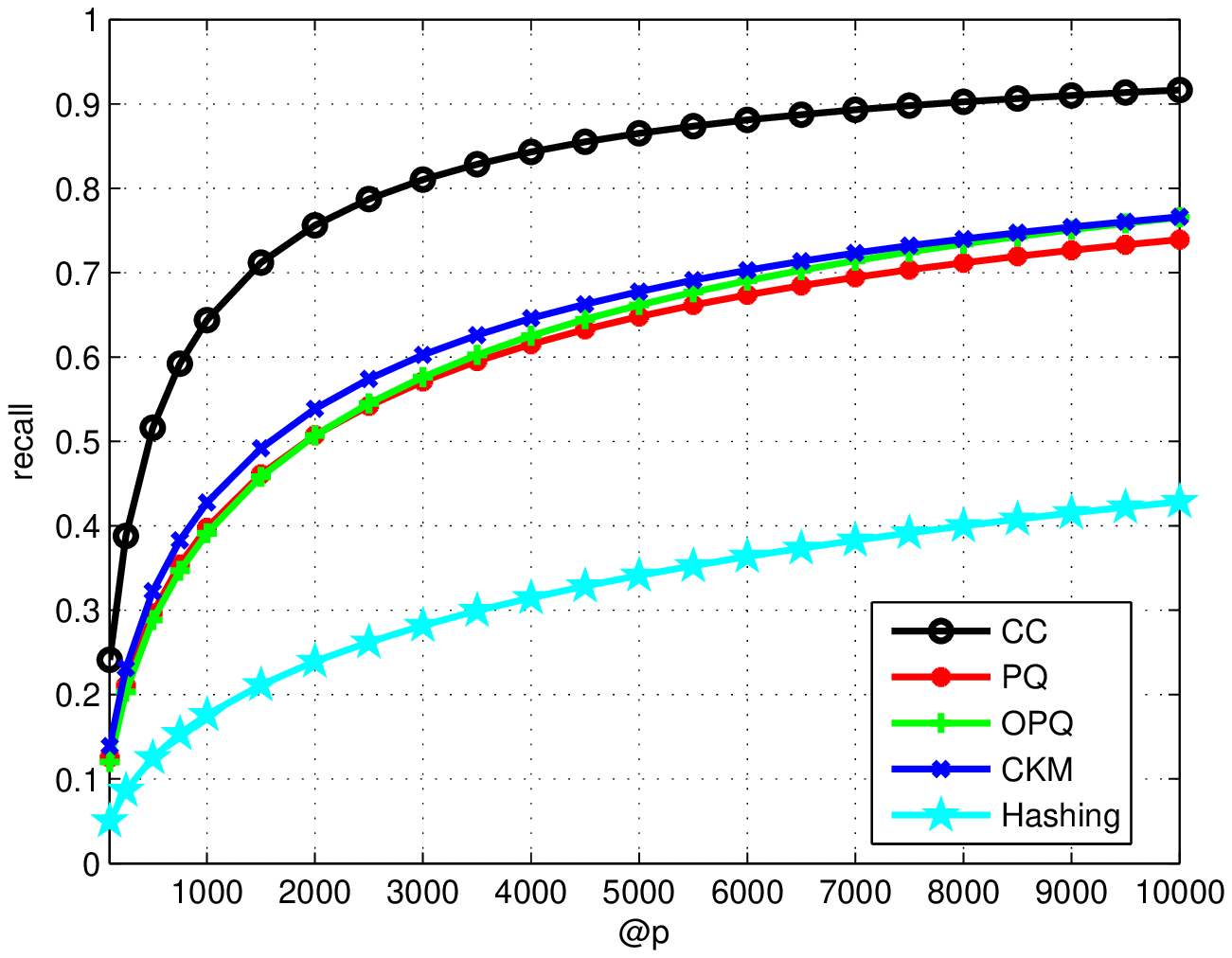}}
\vspace{-0.2cm}
\caption{Performance comparison over SIFT$1B$ with $64$ (top) and $128$ (bottom) bits
for searching (a) $1$-NN, (b) $10$-NNs, and (c) $100$-NNs}
\label{fig:SIFT1BResultsComparison}\vspace{-0.5cm}
\end{figure*}

\begin{figure*}[t]
\centering
\subfigure{\includegraphics[width=.29\textwidth, clip]{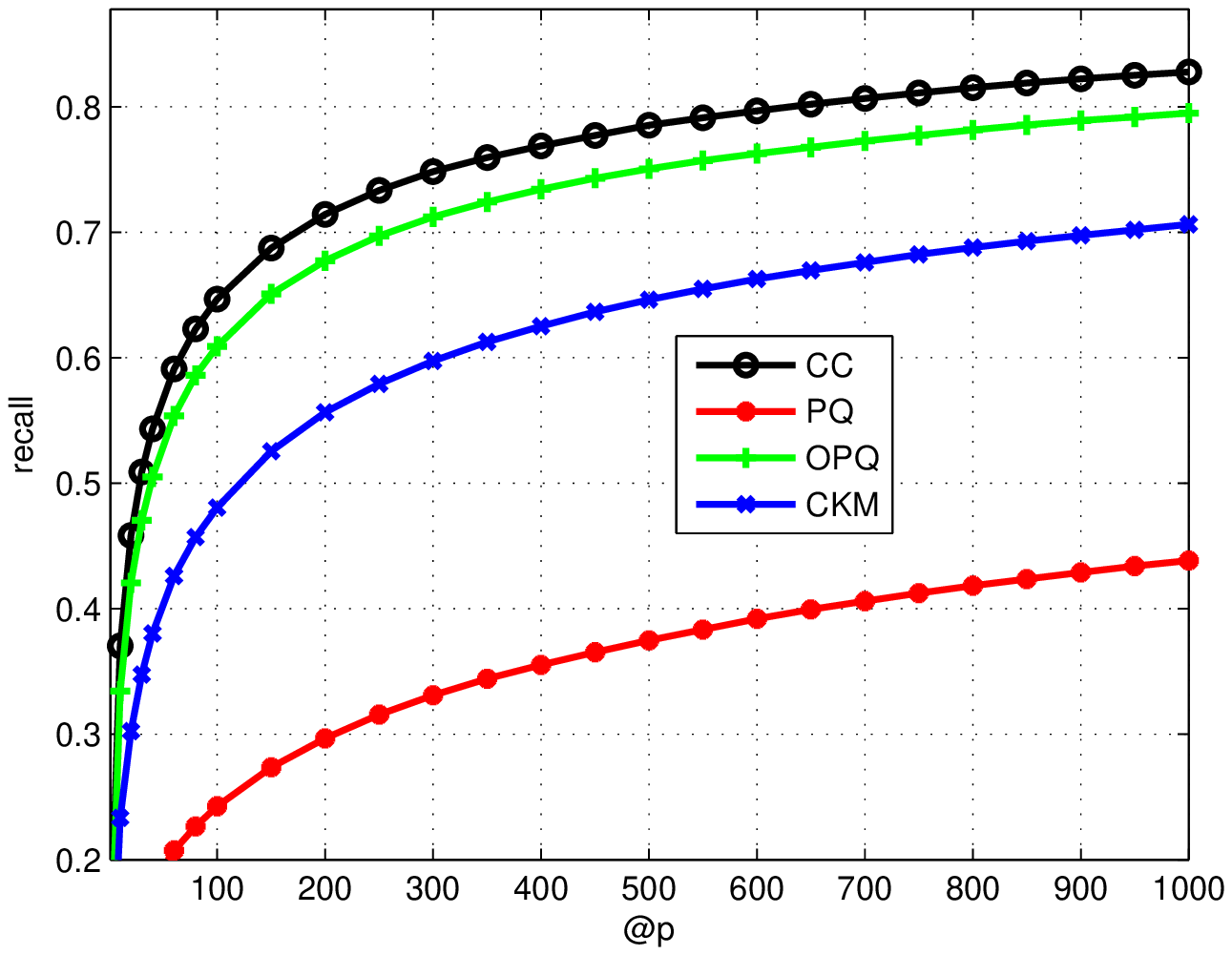}}~~~~~
\subfigure{\includegraphics[width=.29\textwidth, clip]{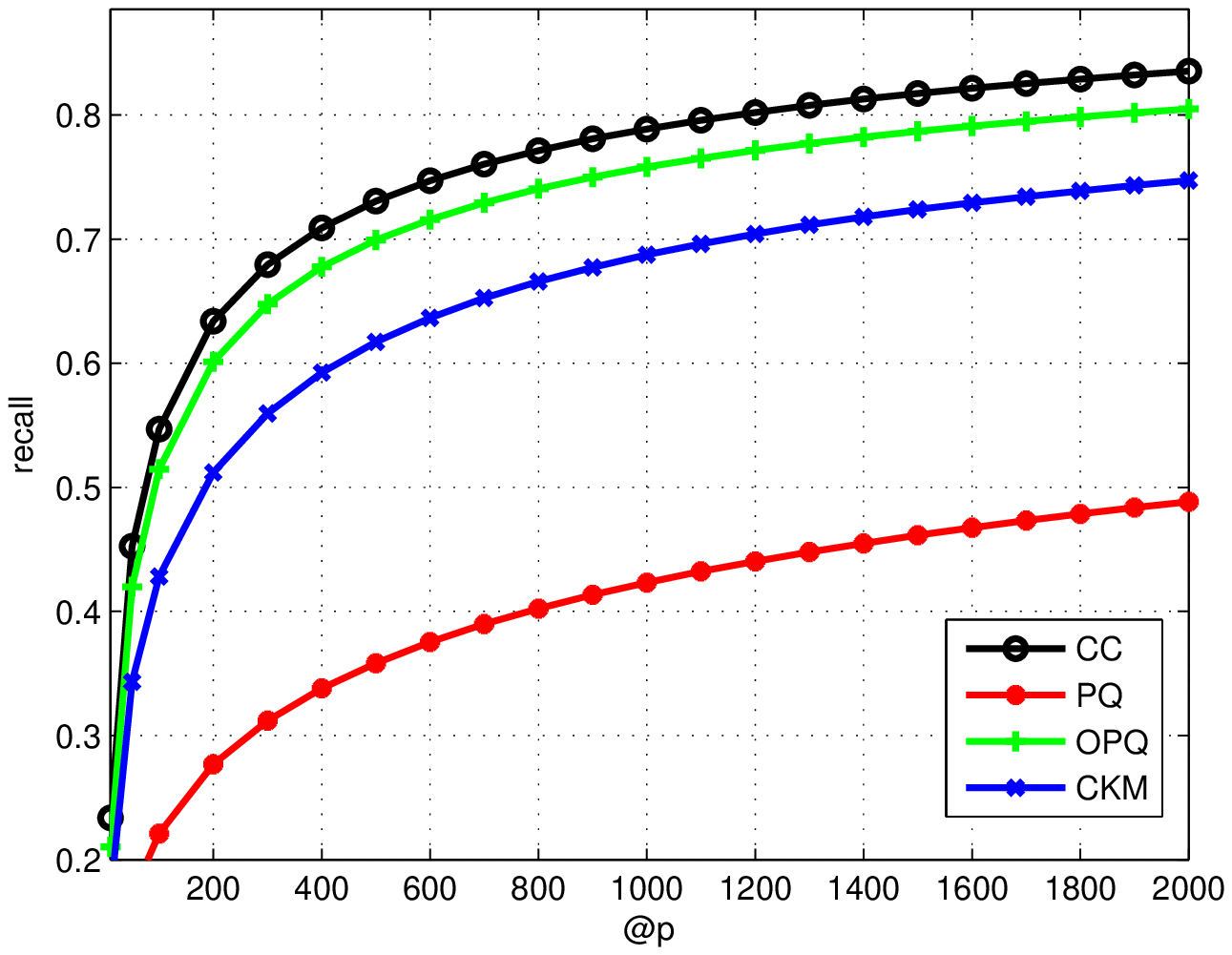}}~~~~~
\subfigure{\includegraphics[width=.29\textwidth, clip]{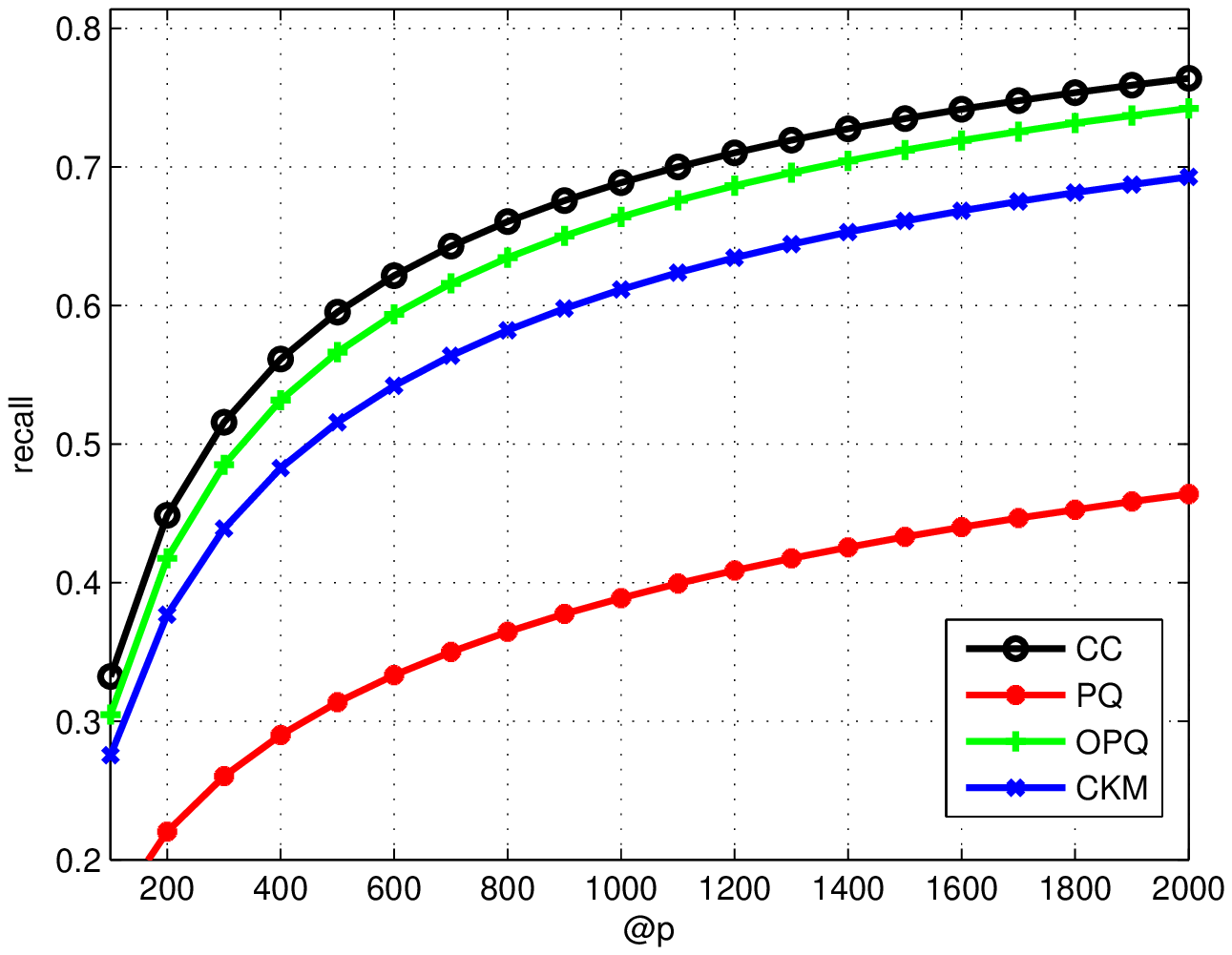}}\\
\vspace{-.2cm}
\subfigure[(a)]{\includegraphics[width=.29\textwidth, clip]{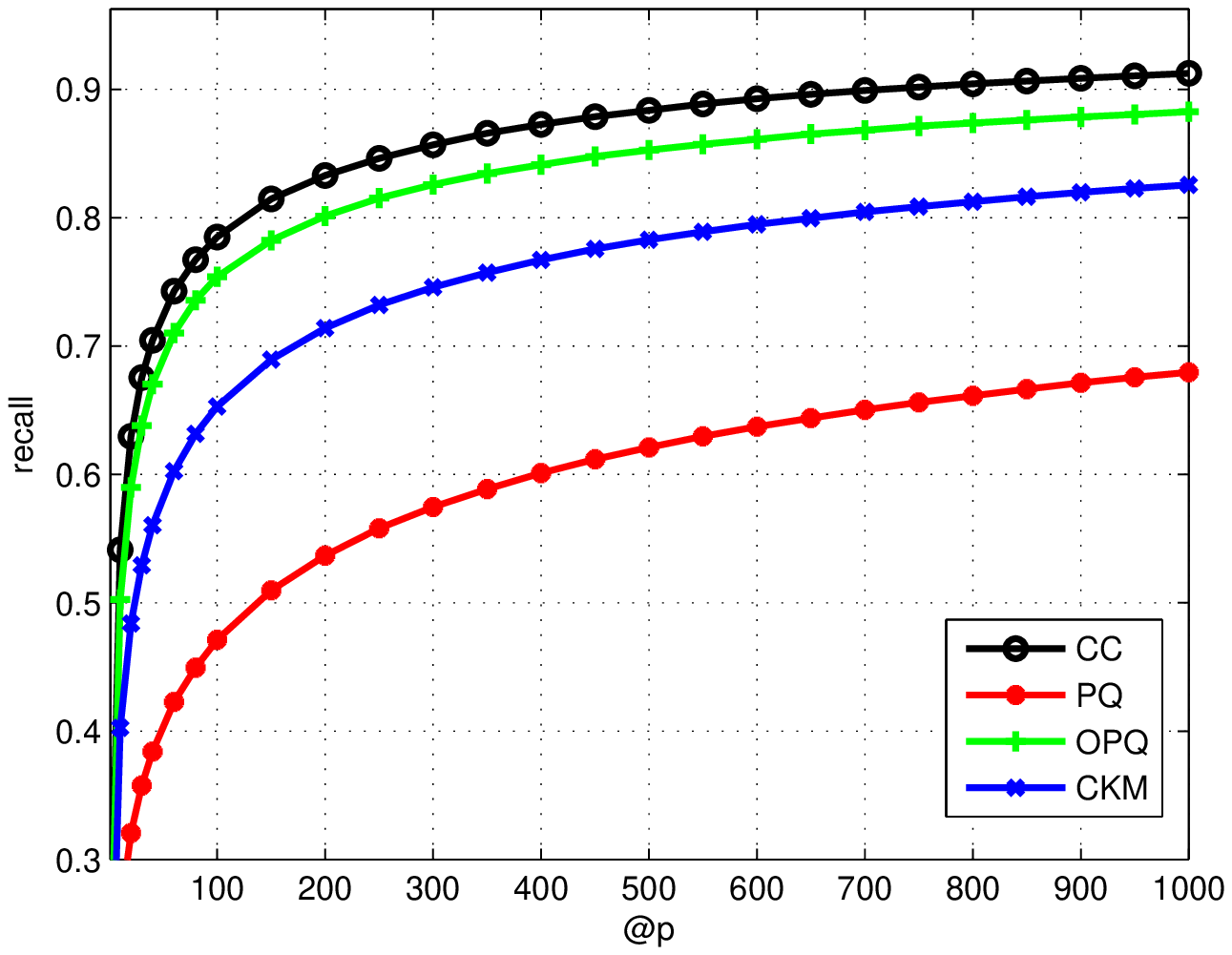}}~~~~~
\subfigure[(b)]{\includegraphics[width=.29\textwidth, clip]{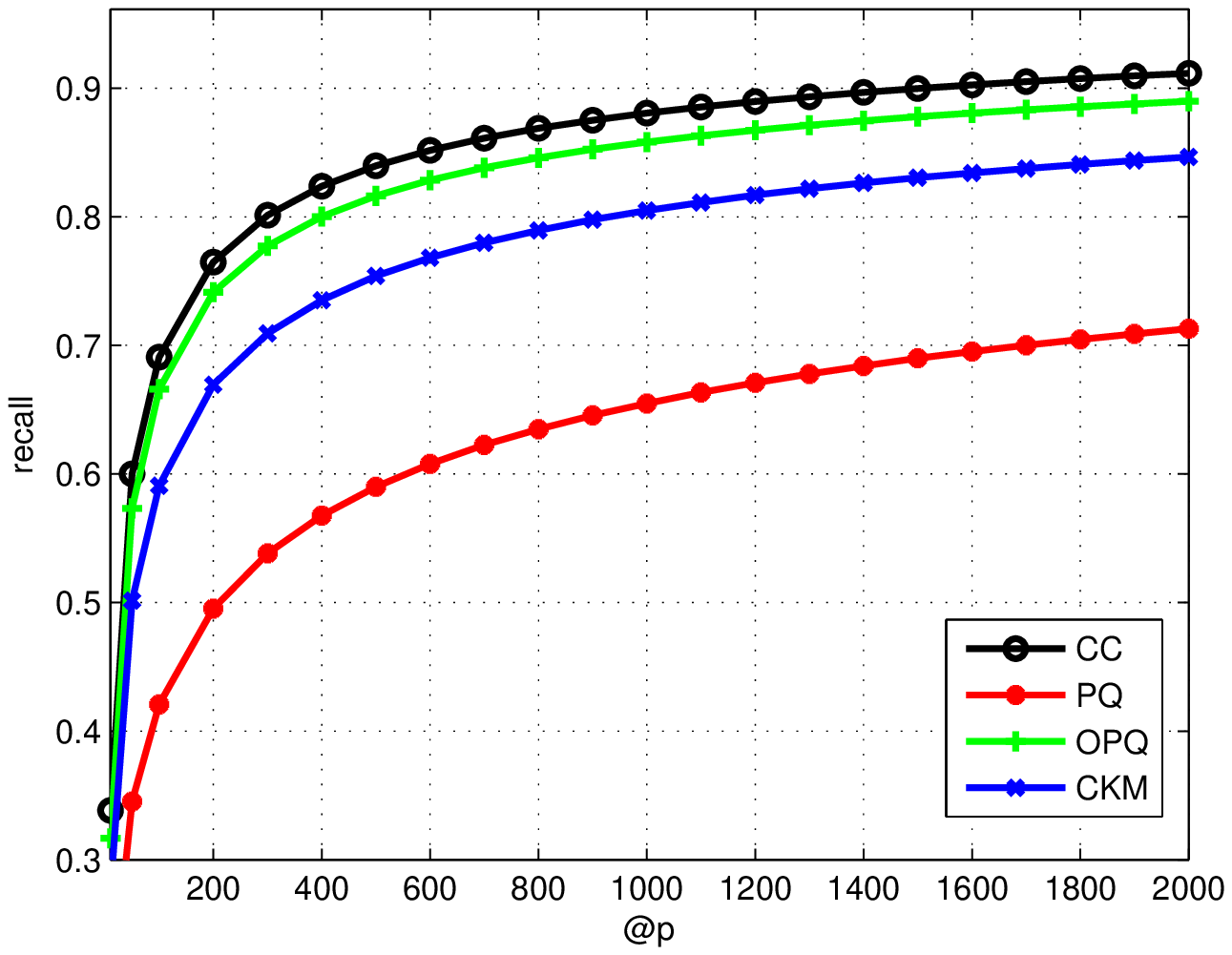}}~~~~~
\subfigure[(c)]{\includegraphics[width=.29\textwidth, clip]{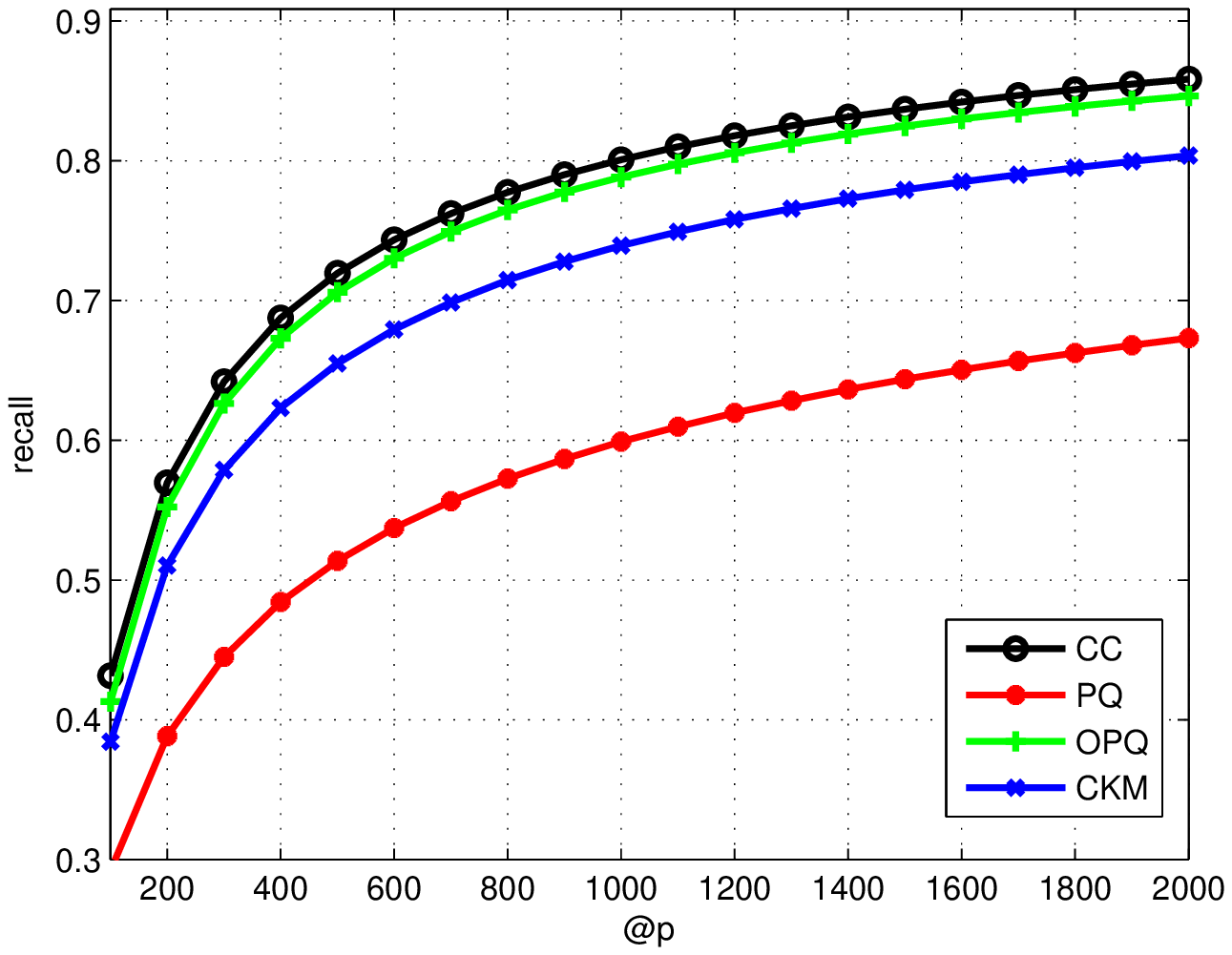}}
\vspace{-0.2cm}
\caption{Performance comparison over LM$1M$ with $64$ (top) and $128$ (bottom) bits
for searching (a) $1$, (b) $10$, and (c) $100$ most relevant linear models}
\label{fig:ClassifierResultsComparison}\vspace{-0.2cm}
\end{figure*}

\begin{figure*}[t]
\centering
\subfigure[(a)]{\includegraphics[width=.24\linewidth, clip]{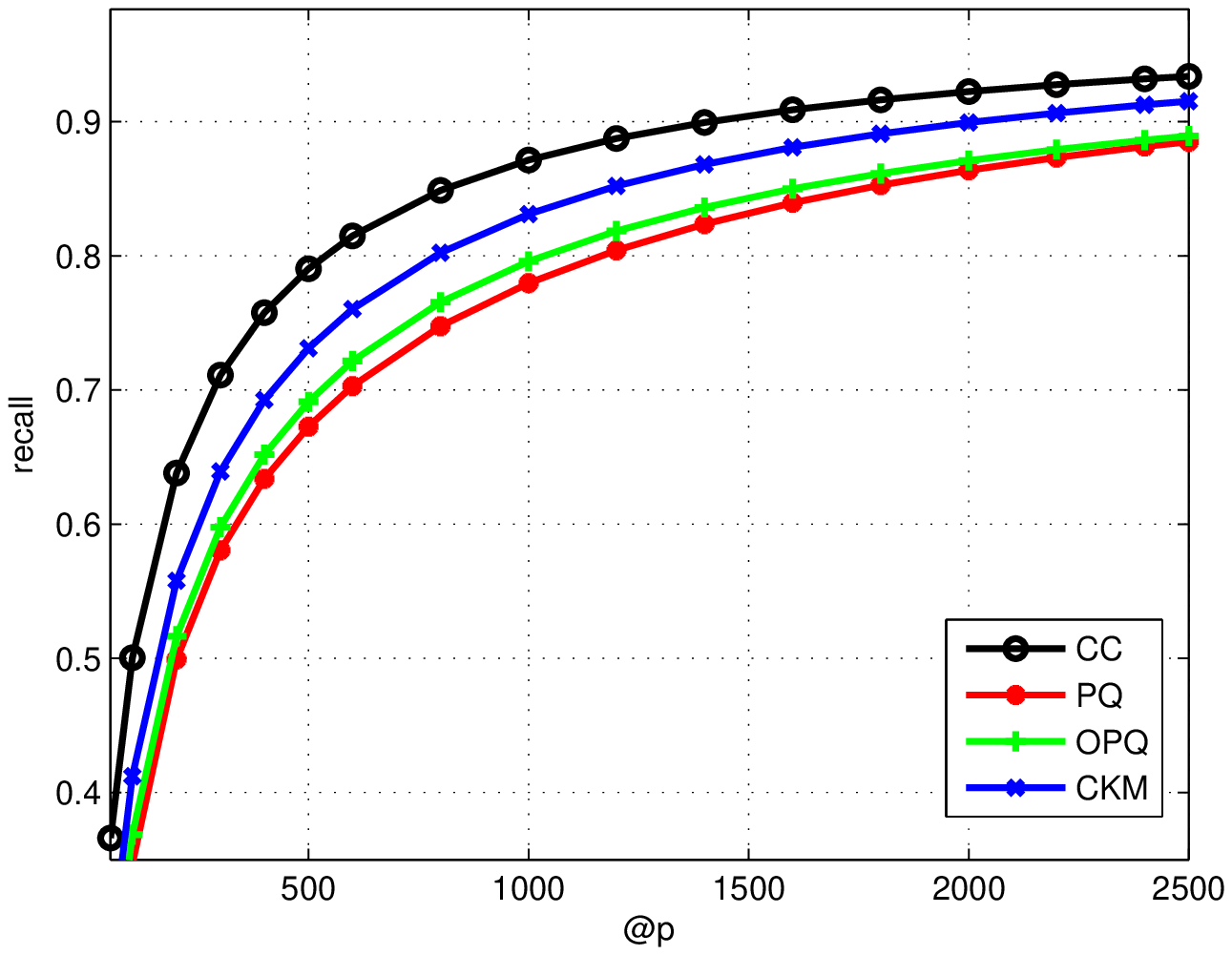}}~
\subfigure[(b)]{\includegraphics[width=.24\linewidth, clip]{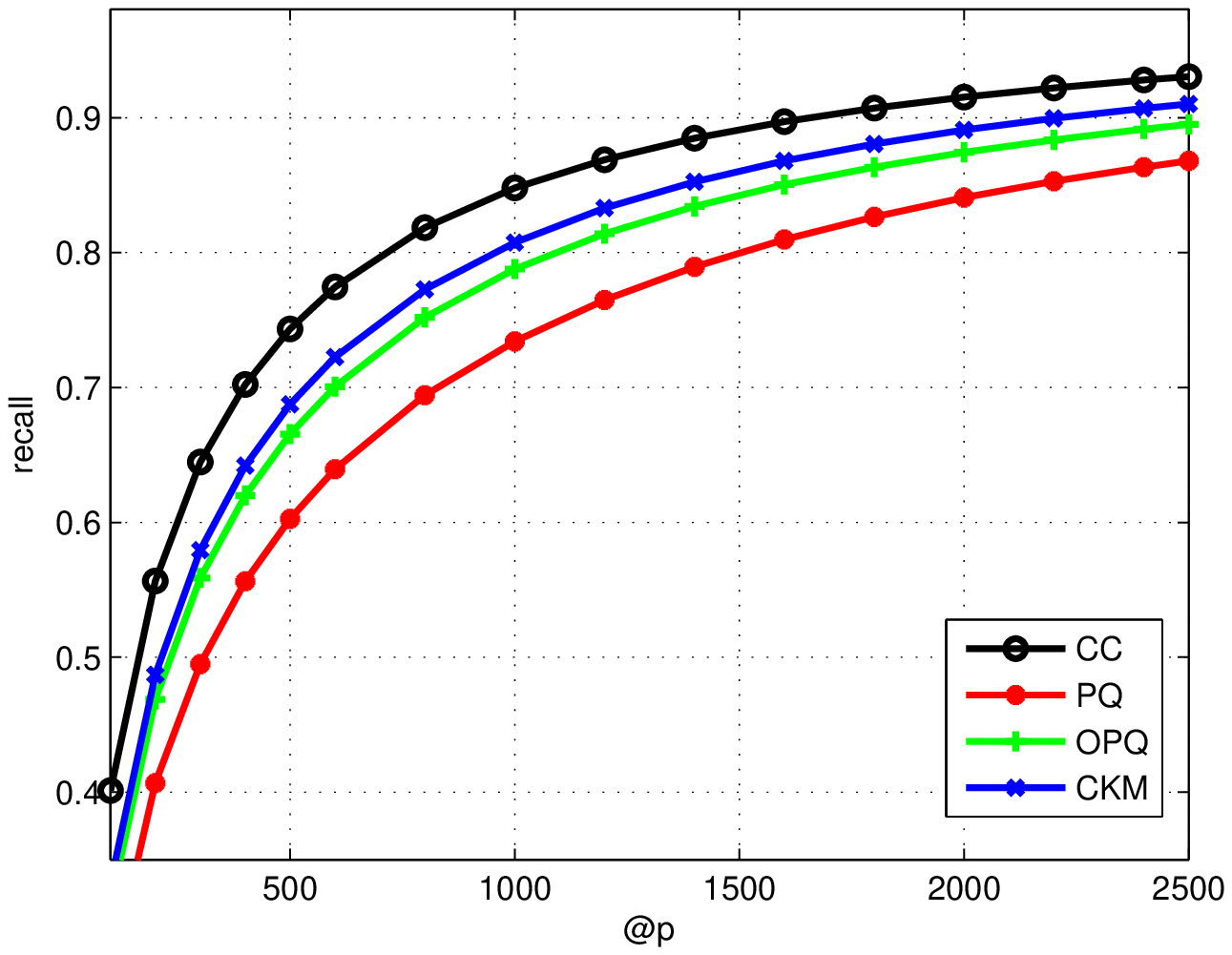}}~
\subfigure[(c)]{\includegraphics[width=.24\linewidth, clip]{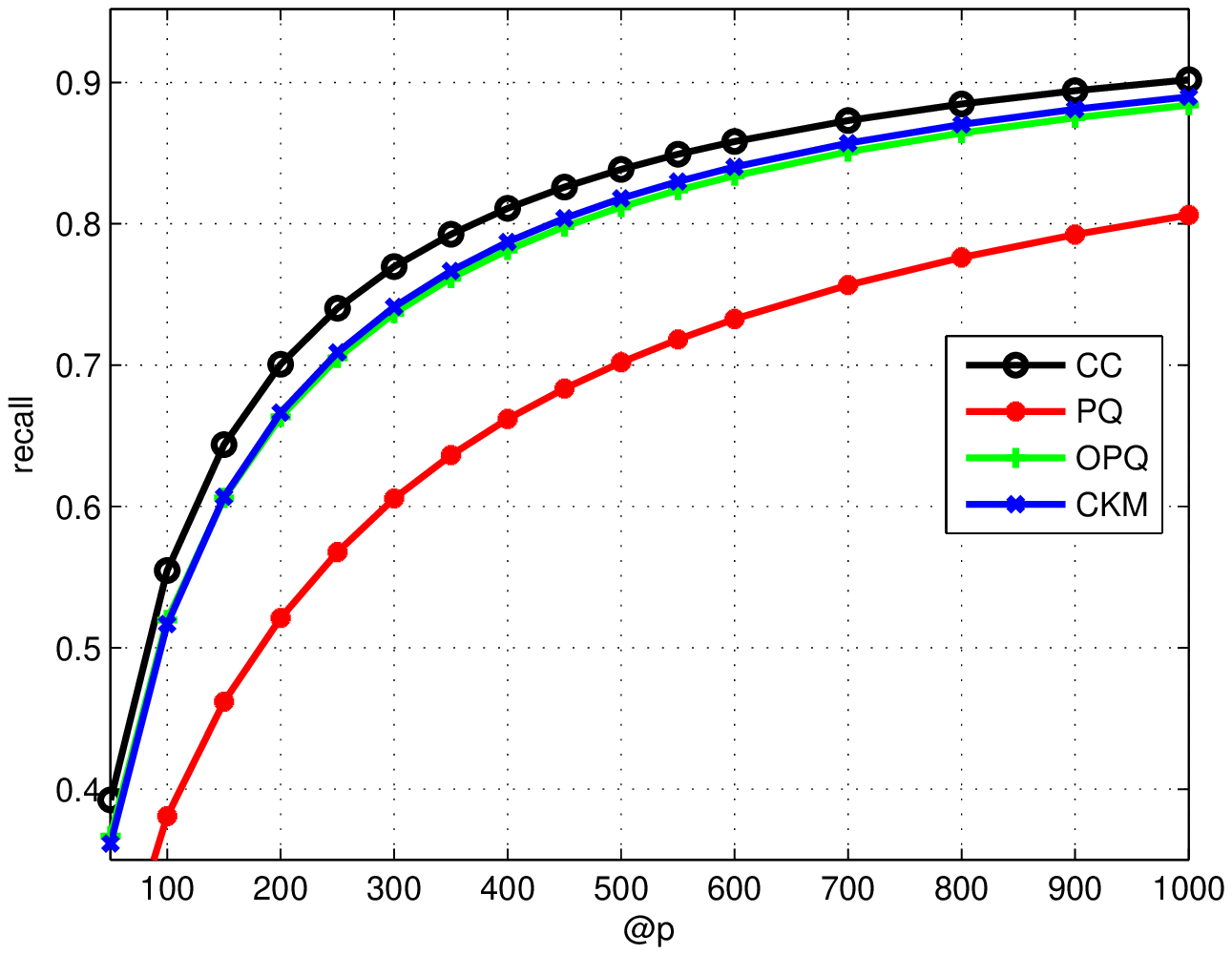}}~
\subfigure[(d)]{\includegraphics[width=.24\linewidth, clip]{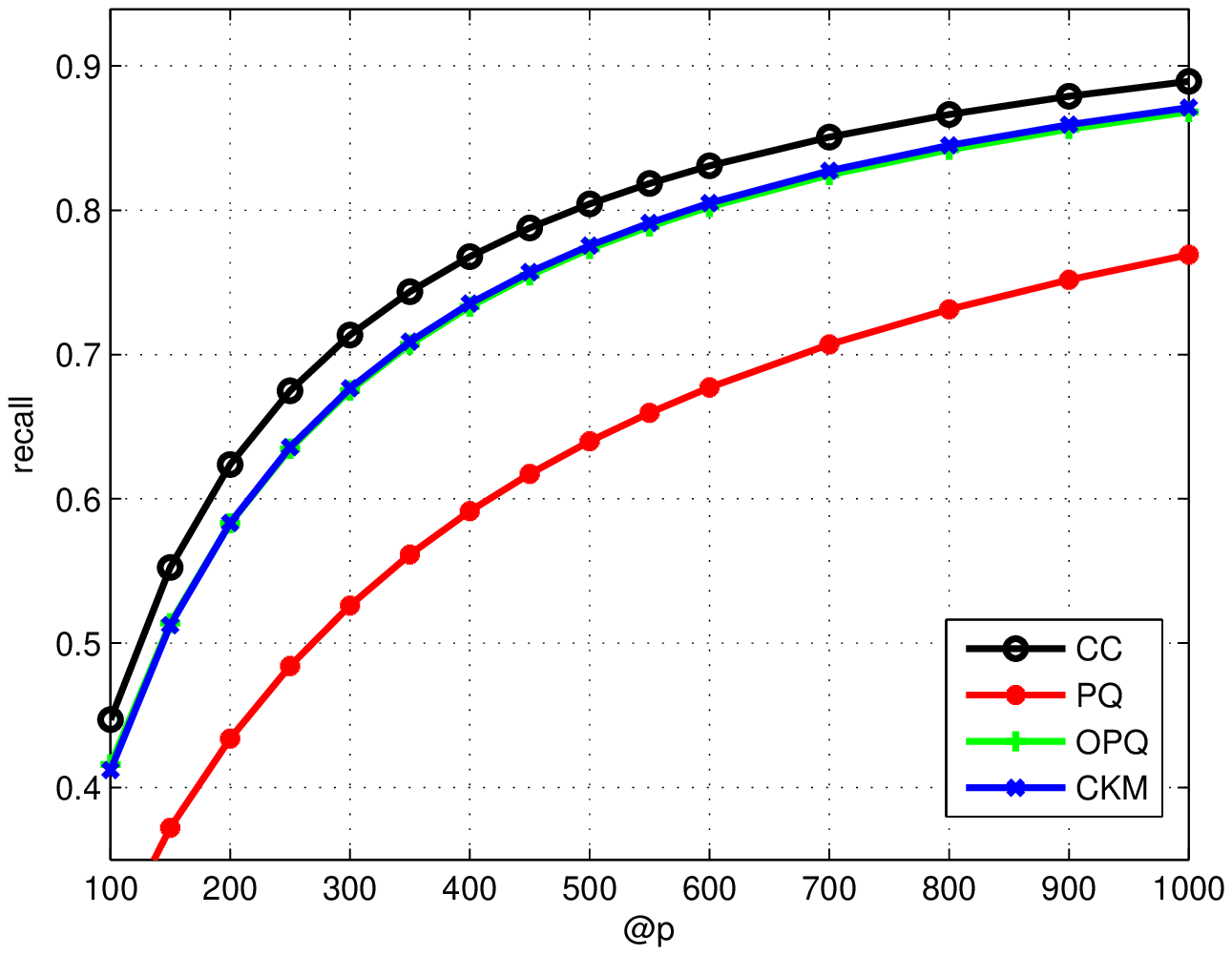}}
\vspace{-0.3cm}
\caption{Performance comparison over Netflix with $32$ ((a) and (b)) and $64$ ((c) and (d)) bits
for searching top $50$ ((a) and (c)) and $100$ ((b) and (d)) users with similar interest}
\label{fig:NetflixResultsComparison}\vspace{-0.3cm}
\end{figure*}

\subsection{Comparison}
We compare our approach,
compositional code with group $M$-selection (abbreviated as CC),
with several compact coding algorithms,
including product quantization (PQ)~\cite{JegouDS11},
the eigenvalue allocation scheme of optimized product quantization (abbreviated as OPQ for convenience)~\cite{GeHK013},
and Cartesian $k$-means~\cite{NorouziF13}
(CKM,
equivalent to optimized product quantization~\cite{GeHK013}).
All the results are achieved
using the same variables of $M$ and $K(=256)$
from the compared algorithms.
We also report the random projection based locality sensitive hashing
(LSH)~\cite{DatarIIM04} that is designed for cosine similarity search
for comparison.

The experimental results over
SIFT$1M$ are shown in~Figure~\ref{fig:SIFT1MResultsComparison}.
One can see that
our approach (CC)
is superior over other algorithms.
Recall$@500$ with $64$ bits for $1$-NN, $10$-NN
and $100$-NN is $20\%$ larger than those from other algorithms.
One can also observe that
the improvement with $128$ bits becomes smaller
than with $64$ bits.
This is because all the algorithms
with more bits result in smaller approximation error
and thus the inner product approximation quality becomes closer.
Figure~\ref{fig:SIFT1BResultsComparison}
shows the results over a very large dataset, SIFT$1B$.
We can see that there are consistent improvements
and our approach achieves above $20\%$ improvement for Recall$@500$ with $64$ bits for $1$-NN.
In comparison with the hashing algorithm,
our algorithm is much better in the case of using the same code length.
Our experiment indicates that the query time cost of our algorithm
is about $1.8$ times of that of the hashing algorithm.
However,
our algorithm using the code of a half length
still outperforms the hashing algorithm.
For example,
one can observed from Figure~\ref{fig:subfig:b}
that
our method using $64$ bits achieves about $1.5$ search accuracy at $p=1000$ compared with hashing using $128$ bits
and in this case the time cost is even smaller than hashing.

The experimental results over LM$1M$
are shown in Figure~\ref{fig:ClassifierResultsComparison}.
In this case,
the inner product similarity search
aims to find the linear models
which the query image fits the best,
equivalently meaning that
the query image
is most relevant to the textual queries
associated with the linear models.
The retrieved linear models can be viewed as soft attributes
that can be applied to image search ranker.
From the respect of large scale classification,
our approach can provide fast prediction
for a large number of categories,
which is a flat approach
rather than hierarchical label trees recently studied
(e.g.~\cite{DengSBL11}).
We present the performance
to show how the approximate search algorithms are close to the exact search algorithm.
From the results shown in Figure~\ref{fig:ClassifierResultsComparison},
we can see that the recall improvement $@300$ with $64$ bits for $1$-NN
is above $3\%$
over the second best,
$14\%$ over the others,
and our approach performs the best.

We also show the performance over the Netflix dataset
in Figure~\ref{fig:NetflixResultsComparison}.
The task in this experiment
is that we retrieves
the similar users
by viewing the rating vector
as the feature of one user,
which can be applied to
mine the films that the query user might be interested
from the films rated by the similar users.
One can see that our approach performs much better
with $32$ bits,
showing the advantage of approach under very small codes,
and the improvement with $64$ bits is a little small,
which might come from that the code of $64$ bits
is already able to well characterize the differences.
One point observed from all the four comparisons
is that our approach consistently performs the best
while no other algorithm always performs the second best.

Last,
we show the advantage
in the potential application
of learning large scale image classifiers,
beyond similarity search.
Image classification
with a large scale
is shown to
achieve state-of-the-art performance
with the use of high-dimensional signatures~\cite{PerronninAHS12,SanchezP11}.
\cite{PerronninAHS12,SanchezP11} show
that data compression is necessary
to support efficient in-RAM training
with stochastic gradient decent (SGD)
as the raw training features
are too large to be loaded into the memory
in normal PCs.

The training process~\cite{PerronninAHS12,SanchezP11}
needs to decompress
the compact code and pass the decompressed version
to the SGD iteration.
It is expected
that the decompressed version is as close as the raw feature as possible.
Thus, we first use the closeness,
i.e., the average feature approximation error
($\operatorname{E}_i[\|\mathbf{x}_i - \bar{\mathbf{x}}_i\|_2^2]$
with $\bar{\mathbf{x}}$ being the decompressed vector),
as a criterion.
The performance is reported
over the $4096$-dimensional fisher vectors~\cite{SanchezP11}
extracted from
the INRIA holidays dataset~\cite{JegouDS08}
that contains $500$ query and $991$ corresponding relevant images,
and the UKbench Recognition Benchmark images~\cite{NisterS06}
that contains $10200$ images.
The conclusions from such two datasets
hold for larger scale datasets.
From the results shown in Tables~\ref{tab:Holidayclassification} and~\ref{tab:UKBclassification},
one can see that
our approach can achieve the best vector approximation.

One popular category of classification algorithms in
large-scale image classification is linear SVM or its variants,
e.g., used in~\cite{PerronninAHS12,SanchezP11}, in which the
training equivalently depends on the inner product
approximation as the dual formulation is based on the kernel
matrix formed by the inner products of the training features.
So we also compare the inner product approximation accuracy
($\operatorname{E}_{ij}[(\mathbf{x}_i^T\mathbf{x}_j -
\bar{\mathbf{x}}_i^T\bar{\mathbf{x}}_j)^2]$) as shown in
Tables~\ref{tab:Holidayclassification}
and~\ref{tab:UKBclassification}.
Here the inner product is
evaluated
in a symmetric way as the training algorithm can only
use the decompressed features. Note that the computation cost
of symmetric inner product, $O(M^2)$, does not matter because
the SGD algorithm does not really compute the inner product. In
addition, we also show that such approximate (symmetric) inner
product similarities is also superior in preserving the
semantic similarity by evaluating the search performance in
terms of mean average precision (MAP)~\cite{JegouDS08} for the
holiday dataset and
score~\cite{NisterS06}
for the UKBench
dataset.

\setlength{\tabcolsep}{4pt}
\begin{table}[!t]
\begin{center}
\caption{Performance comparison in the application of data
compression using short codes over the holidays dataset.
VAE = vector approximate error.
IPAE = inner product approximation
error}
\footnotesize
\label{tab:Holidayclassification}
\begin{tabular}{ l|lll |lll}
   \hline
   & \multicolumn{3}{c|}{$32$ bits} & \multicolumn{3}{c}{$64$ bits} \\
   \hline
   & VAE & IPAE & MAP & VAE & IPAE & MAP  \\
   \hline
   PQ & $0.4326$ & $0.0057$ & $0.4179$ & $0.3775$ & $0.0044$ & $0.4740$ \\
   OPQ & $0.4014$ & $0.0049$ & $0.4339$ & $0.3221$ & $0.0030$ & $0.4787$\\
   CKM & $0.3081$ & $0.0054$ & $0.4843$ & $0.2279$ & $0.0039$ & $0.5251$\\
   CC & $0.1462$ & $0.0020$  &  $0.5245$ & $0.0139$ & $.00002$  & $0.6360$ \\
   \hline
\end{tabular}
\end{center}
\vspace{-.5cm}
\end{table}
\setlength{\tabcolsep}{1.4pt}

\setlength{\tabcolsep}{4pt}
\begin{table}[!h]
\begin{center}
\caption{Performance comparison in the application of data
compression using short codes over the UKBench dataset}
\label{tab:UKBclassification}
\footnotesize
\begin{tabular}{ l|lll |lll}
   \hline
   & \multicolumn{3}{c|}{$32$ bits} & \multicolumn{3}{c}{$64$ bits} \\
   \hline
   & VAE & IPAE & score & VAE & IPAE & score  \\
   \hline
   PQ & $0.4629$ & $0.0075$ & $1.871$ & $0.4120$ & $0.0054$ &  $2.180$  \\
   OPQ & $0.3982$ & $0.0050$ & $1.790$ & $0.3232$ & $0.0025$ & $2.193$ \\
   CKM& $0.3767$ & $0.0059$ & $2.064$ & $0.3079$ & $0.0037$ & $2.382$ \\
   CC & $0.3076$ & $0.0038$  & $2.152$ & $0.2261$ & $0.0021$  & $2.551$ \\
   \hline
\end{tabular}
\vspace{-.5cm}
\end{center}
\end{table}
\setlength{\tabcolsep}{1.4pt}

\section{Conclusion}
This paper
studies the approximate inner product similarity search problem
and introduces
a compact code approach,
compositional code using group $M$-selection.
Vector approximation of our approach
is more accurate,
without increasing the code length.
The similarity search is efficient
as
evaluating the approximate inner product
between a query and a compact code
is computationally cheap.
In the future,
we will generalize our approach
for search with Euclidean distance
and other similarity measures.

\section{Appendix: Proof}
We rewrite Property~\ref{property:innerproductbound} presented in the main paper
in the following
and then give the proof\footnote{There is a similar theorem (Theorem $3.1$) in~\cite{RamG12}
showing the maximum value of the approximate inner product
in the same condition.
Differently, we provide the upper-bound of the approximation error
(including both maximum and minimum values of the approximate inner product)
and present a more succinct proof.}.

\begin{Property}
\label{property:innerproductbound}
Given a data vector $\mathbf{p}$
and a query vector $\mathbf{q}$,
if the distance between $\mathbf{p}$
and its approximation $\bar{\mathbf{p}}$
is not larger than $r$,
$\|\mathbf{p} - \bar{\mathbf{p}}\|_2 \leqslant r$,
then the absolute difference between
the true inner product
and the approximate inner product
is upper-bounded:
\begin{align}
|\texttt{<} \mathbf{q}, \mathbf{p} \texttt{>} - \texttt{<} \mathbf{q}, \bar{\mathbf{p}} \texttt{>}|
\leqslant r \|\mathbf{q}\|_2.
\end{align}
\end{Property}
\begin{proof}
The proof is simple
and given as follows.
Let $\bar{\mathbf{p}} = \mathbf{p} + \boldsymbol{\delta}$.
By definition, we have $\|\boldsymbol{\delta}\|_2 \leq r$.
Look at the absolute value of the inner product approximation error,
\begin{align}
&\|\texttt{<} \mathbf{q}, \bar{\mathbf{p}} \texttt{>} - \texttt{<}\mathbf{q}, \mathbf{p}\texttt{>}\| \\
=~& \|\texttt{<}\mathbf{q}, \mathbf{p} + \boldsymbol{\delta}\texttt{>} - \texttt{<}\mathbf{q}, \mathbf{p}\texttt{>} \| \\
=~& \|(\texttt{<}\mathbf{q}, \mathbf{p}\texttt{>} + \texttt{<}\mathbf{q}, \boldsymbol{\delta}\texttt{>})
- \texttt{<}\mathbf{q}, \mathbf{p}\texttt{>} \|~\text{\emph {(by the distributive property)}}\\
=~& \|\texttt{<}\mathbf{q}, \boldsymbol{\delta}\texttt{>}\| \\
\leq~&\| \boldsymbol{\delta} \|_2 \| \mathbf{q} \|_2 \\
\leq~& r\| \mathbf{q} \|_2.
\end{align}
Thus, the approximation error is upper-bounded by
$r\| \mathbf{q} \|_2$.
\end{proof}

\section{Appendix: Analysis}
\subsection{Connection to Product Quantization and Cartesian $K$-means}
The idea of product quantization~\cite{JegouDS11} is
to decompose the space into a Cartesian product
of $M$ low dimensional subspaces
and to quantize each subspace separately.
A vector $\mathbf{x}$ is then
decomposed into $M$ subvectors,
$\mathbf{x}^1, \cdots, \mathbf{x}^M$.
Let the quantization dictionaries
over the $M$ subspaces
be $\mathcal{C}_1, \mathcal{C}_2, \cdots, \mathcal{C}_M$
with $\mathcal{C}_m$ being
a set of centers $\{\mathbf{c}_{m1}, \cdots, \mathbf{c}_{mK}\}$.
A vector $\mathbf{x}$ is represented
by the concatenation of $M$ centers,
$[\mathbf{c}_{1k_1^*}^T \mathbf{c}_{2k_2^*}^T
\cdots  \mathbf{c}_{mk_m^*}^T
\cdots  \mathbf{c}_{Mk_M^*}^T]^T$
each of which $\mathbf{c}_{mk_m^*}$ is the one nearest
to $\mathbf{x}^m$
in the $m$-th quantization dictionary, respectively.

Rewrite each center $\mathbf{c}_{mk}$
as a $d$-dimensional vector $\tilde{\mathbf{c}}_{mk}$
so that
$\tilde{\mathbf{c}}_{mk} = [\boldsymbol{0}^T, \cdots, (\mathbf{c}_{mk})^T, \cdots, \boldsymbol{0}^T]^T$,
i.e., all entries are zero
except that the part corresponding to the $m$th subspace
is equal to $\mathbf{c}_{mk}$.
The approximation of a vector $\mathbf{x}$
using the concatenation
$\bar{\mathbf{x}} = [\mathbf{c}_{1k_1^*}^T~\mathbf{c}_{2k_2^*}^T \cdots \mathbf{c}_{Mk_M^*}^T]^T$
is equivalent to
the composition
$\bar{\mathbf{x}} = \sum_{m=1}^M \tilde{\mathbf{c}}_{mk_m^*}$.

Cartesian $k$-means~\cite{NorouziF13} extends the subspace decomposition
by performing a rotation $\mathbf{R}$
($\mathbf{R}^T\mathbf{x}$)
and then the product quantization
in the rotated space,
where the rotation and the subquantizers
are jointly optimized.
Similar to product quantization,
the vector approximation in Cartesian $k$-means
is equivalent to
$\bar{\mathbf{x}} = \sum_{m=1}^M \mathbf{R} \tilde{\mathbf{c}}_{mk_m^*}
= \sum_{m=1}^M \tilde{\mathbf{c}}'_{mk_m*}$.

From the above analysis,
the vector approximation approach using product quantization
and Cartesian $k$-means
can be viewed as a constrained version
of our approach,
each subquantizer corresponds to
a source dictionary in our approach
and each source dictionary lies in a different subspace
with the same dimension
in the case that each subspace is full-ranked.
If some subspaces are not full-ranked,
the equivalence still holds.

\subsection{Relation to Order Permutation}
Order permutation~\cite{ChavezFN08}
is an index algorithm
for approximate nearest neighbor search.
It aims to predict closeness
between data vectors according to how they order their distances
towards a distinguished set of anchor vectors
(such as the $k$-means centers $\mathcal{C}$).
Each data vector sorts the anchor objects from
closest to farthest to it,
i.e., a permutation of anchor objects,
$\{\mathbf{c}_{i_1^*}, \mathbf{c}_{i_2^*}, \cdots, \mathbf{c}_{i_t^*}\}$
($t \leqslant K$),
and the similarity between orders
is used as a predictor of the
closeness between the corresponding elements.

In contrast,
our approach finds a partial permutation of
the source dictionary
and uses their composition (rather than the order) to approximate
the data vector,
yielding a compact code representation.
In the query stage,
the distance of the query to the approximated data vector,
in a fast way by the efficient
table-lookup and addition operations,
is computed
to approximate the closeness.

\subsection{Connection to Sparse Coding}
The aim of sparse coding is
to find a set of $K$ (over-complete) basis vectors
$\{\boldsymbol{\phi}_k\}$
such that
a data vector $\mathbf{x}$ as a linear combination
of these basis vectors:
$\mathbf{x} = \sum_{k=1}^K \alpha_k\boldsymbol{\phi}_k$,
where there are few non-zero coefficients
in the coefficient vector $\boldsymbol{\alpha} = [\alpha_1~\alpha_2\cdots \alpha_K]^T$,
i.e., $\|\boldsymbol{\alpha}\|_0$ is small.
The proposed $M$-combination scheme
can be viewed as a special sparse coding approach
in which the coefficients can be only valued as $0$ or $1$
and the sparsity is fixed, $\|\boldsymbol{\alpha}\|_0 = M$.

Coding with group sparsity~\cite{YuanL06}
is an extension of sparse coding,
in which the coefficients
$\{\boldsymbol{\phi}_k\}$ are divided into several groups
and the sparsity constraints are imposed in each group separately.
The proposed $M$-selection and group $M$-selection schemes
can be regarded as a special coding approach
with group sparsity,
where the coefficients that can be only
valued by $0$ or $1$
are divided into $M$ groups
and for each group the non-sparsity degree is $1$.

\subsection{Time Complexity}
In the main paper,
we have shown that
the time complexity of dictionary updating is $O(NMd + d (MK)^2 + NM^2 + (MK)^3)$
and the time complexity of code computation (group $M$-selection updating)
is $O(NM^2Kd)$.
The following gives detailed analysis on the time complexity.

The dictionary is updated in our algorithm
by computing the closed-form solution:
$\mathbf{D} = \mathbf{X}\mathbf{B}^T(\mathbf{B}\mathbf{B}^T)^{-1}$.
The computation consists of
(1) the matrix multiplication operation: $\mathbf{X}\mathbf{B}^T$ ($=\mathbf{E}$),
(2) the matrix multiplication operation: $\mathbf{B}\mathbf{B}^T$ ($=\mathbf{Q}$),
(3) the inverse operation: $\mathbf{Q}^{-1}$ ($=\mathbf{R}$),
and (4) the multiplication operation: $\mathbf{E}\mathbf{R}$.
Note that $\mathbf{X}$ is a matrix of $d\times N$,
$\mathbf{B} = [\mathbf{b}_1\cdots\mathbf{b}_N]$
and each $\mathbf{b}_n$ is a $MK$-dimensional vector
with only $M$ entries being $1$ and all the others being $0$.
It can be easily shown that
$\mathbf{E}$ is of size $d \times MK$,
$\mathbf{Q}$ is of size $MK \times MK$,
and $\mathbf{R}$ is of size $MK \times MK$.
Step (1) takes $O(NMd)$ due to the sparse matrix $\mathbf{B}$.
Step (2) can be transformed to $\mathbf{B}\mathbf{B}^T = \sum_{n=1}^N\mathbf{b}_n\mathbf{b}_n^T$.
Because $\mathbf{b}_n$ is a sparse vector,
$\mathbf{b}_n\mathbf{b}_n^T$ takes $O(M^2)$ instead of $O(M^2K^2)$.
Step (3) takes $O((MK)^3)$ ,
and step (4) takes $O(d(MK)^2)$.
In summary, the whole time complexity is $O(NMd + d (MK)^2 + NM^2 + (MK)^3)$.

The code (group $M$-selection) is updated
by optimizing $\mathbf{b}_n$ separately.
Each $\mathbf{b}_n$ is computed
as $\min_{\mathbf{b}_n} = \|\mathbf{x}_n -\mathbf{D}\mathbf{b}_n\|_2^2$.
We solve it by a greedy algorithm,
performing $M$ $1$-selection optimizations over the $M$ source dictionaries
in the best-first manner.
The $m$-th optimization involves selecting the best $1$-selection over $(M-m-1)$ source dictionaries,
each of which contains $K$ $d$-dimensional exemplar vectors.
It costs $O\left((M-m-1)(Kd)\right)$ to select the best exemplar vector.
There are $M$ optimizations to be performed,
thus the cost of updating $\mathbf{b}_n$
is $\sum_{m=1}^MO\left((M-m-1)(Kd)\right)=O(M^2Kd)$.
In summary, the time complexity of updating all $\mathbf{b}_n$
is $O(NM^2Kd)$.

\section{Appendix: More Experimental Results}
In the main paper,
we show the inner product similarity search performance
over four datasets.
Here we report the average approximation error using the composition of the selected vectors as the vector approximation
over the database vectors,
and the average inner production approximation error using the approximated vector
between the query vector and the nearest $100$ database vector
as shown in Tables~\ref{tab:64bits} and~\ref{tab:128bits}.
One can observe that the average vector approximation error of our approach is the smallest
and the inner product approximation error is also the smallest.
This gives another evidence
that our approach can achieve the best similarity search performance.

\begin{table*}[t]
\centering
\footnotesize
\caption{The average vector approximation error for the database vectors,
and the average inner product approximation error
between the query vector and the nearest $100$ database vectors
using $64$ bits.
VAE = vector approximate error.
IPAE = inner product approximation error.}
\label{tab:64bits}
\begin{tabular}{c||c|c|c|c|c|c|c|c}
\hline
& \multicolumn{2}{|c|} {SIFT1M} & \multicolumn{2}{|c|} {SIFT1B} & \multicolumn{2}{|c|} {Netflix} & \multicolumn{2}{|c} {LinerModels}\\
\hline
& VAE ($10^{4}$) & IPAE ($10^{2}$) & VAE ($10^{4}$) & IPAE ($10^{3}$) & VAE ($10^{2}$) & IPAE  & VAE ($10^{-3}$)& IPAE ($10^{-3}$) \\
\hline
PQ & $2.319$ & $8.915$ &  $2.540$ & $1.707$ & $8.588$ & $166.03$ & $4.999$ & $4.755$ \\
\hline
OPQ & $2.842$ & $8.238$ & $3.282$  & $2.154$ & $7.091$ & $52.58$ & $4.364$ & $1.957$ \\
\hline
CKM & $2.134$ & $6.835$ &  $2.346$ & $1.318$ & $6.273$ & $68.78$ & $4.278$ & $2.312$ \\
\hline
CC & $1.626$ & $2.148$ & $1.773$ & $0.460$ & $6.014$ & $48.31$ & $3.797$ & $1.724$ \\
\hline
\end{tabular}
\end{table*}

\begin{table*}[t]
\centering
\caption{The average vector approximation error for the database vectors,
and the average inner product approximation error
between the query vector and the nearest $100$ database vectors
using $128$ bits.}
\label{tab:128bits}
\footnotesize
{
\begin{tabular}{c||c|c|c|c|c|c|c|c}
\hline
& \multicolumn{2}{|c|} {SIFT1M} & \multicolumn{2}{|c|} {SIFT1B} & \multicolumn{2}{|c|} {Netflix} & \multicolumn{2}{|c} {LinerModels}\\
\hline
& VAE ($10^4$) & IPAE ($10^2$) & VAE ($10^4$)& IPAE ($10^2$)& VAE ($10^2$)& IPAE & VAE ($10^{-3}$) & IPAE ($10^{-3}$)\\
\hline
PQ & $1.038$ & $3.191$ & $1.070$ &  $5.075$ & $7.945$ & $157.02$ & $3.235$ & $2.956$ \\
\hline
OPQ & $1.468$ & $2.398$ & $1.620$ &  $5.610$ & $6.031$ & $60.24$ & $2.760$ & $0.984$ \\
\hline
CKM & $0.992$ & $2.768$ & $1.047$ &  $4.269$ & $4.908$ & $49.25$ & $2.775$ & $1.336$ \\
\hline
CC & $0.797$ & $0.868$ & $0.850$ &  $1.252$ & $4.707$ & $75.23$ & $2.534$ & $0.926$ \\
\hline
\end{tabular}
}
\end{table*}

\section{Appendix: Future Work}

\noindent\textbf{Adaptation to cosine similarity search.}
Inner product is equivalent to cosine similarity
in the case that the database vectors
are of the same $L_2$ norm.
Our approach
finds the optimal composition,
however,
without making
the composition vector keep the same norm.
In the future,
we will study the way
of approximating the vector
with maintaining the $L_2$ norm,
e.g. extending spherical $k$-means clustering.

\noindent\textbf{Extension to similarity search under Euclidean distance.}
The experiments show
that using the compact codes learnt from our approach
for Euclidean distance based similarity search
achieves better search accuracy
than product quantization and Cartesian $k$-means.
Because the distributive property
with respect to the Euclidean distance operation
over the addition operation does not hold
($d^2(\mathbf{q}, \sum_{m=1}^M\mathbf{c}_{mk_m})
\neq \sum_{m=1}^M d^2(\mathbf{q}, \mathbf{c}_{mk_m}) $),
the general time cost
of evaluating the approximate Euclidean distance
using the codes produced from our approach
is $\Theta(M^2)$,
which is a little large.
If the $M$ source dictionaries
(i.e., the $M$ subspaces spanned by the $M$ source dictionaries)
are mutually orthogonal
($\texttt{<}\mathbf{c}_{si}, \mathbf{c}_{rj}\texttt{>}= 0,
\forall s \neq r, \forall i, j$.),
the time cost is reduced to $\Theta(M)$
with the constant coefficient $2$
because $\|\mathbf{q} - \sum_{m=1}^M\mathbf{c}_{mk_m})\|_2^2 =
\|\mathbf{q}\|_2^2 + \sum_{m=1}^M \mathbf{c}^2_{mk_m}
- 2 \times \mathbf{q}^T(\sum_{m=1}^M\mathbf{c}_{mk_m})$.

However, we have the following equation:
\begin{align}
& \|\mathbf{q} - \sum_{m=1}^M\mathbf{c}_{mk_m}\|_2^2 \\
=~& (\mathbf{q} - \sum_{m=1}^M\mathbf{c}_{mk_m})^T(\mathbf{q} - \sum_{m=1}^M\mathbf{c}_{mk_m}) \\
&~\text{\emph {(from orthogonality constraints between the items of different dictionaries)}}\\
=~& \mathbf{q}^T\mathbf{q} - 2  \mathbf{q}^T(\sum_{m=1}^M\mathbf{c}_{mk_m}) + \sum_{m=1}^M \mathbf{c}^T_{mk_m}\mathbf{c}_{mk_m}\\
=~& \mathbf{q}^T\mathbf{q} - 2  \mathbf{q}^T(\sum_{m=1}^M\mathbf{c}_{mk_m}) + \sum_{m=1}^M \mathbf{c}^T_{mk_m}\mathbf{c}_{mk_m}+(M-1)\mathbf{q}^T\mathbf{q}-(M-1)\mathbf{q}^T\mathbf{q} \\
=~&
\sum_{m=1}^M(\mathbf{q}^T\mathbf{q} - 2\mathbf{q}^T\mathbf{c}_{mk_m} + \mathbf{c}^T_{mk_m}\mathbf{c}_{mk_m})-(M-1)\mathbf{q}^T\mathbf{q} \\
=~&
\sum_{m=1}^M\|\mathbf{q} - \mathbf{c}_{mk_m}\|_2^2 - (M-1)\|\mathbf{q}\|_2^2.
\end{align}
The above equations show that,
given a query,
it is enough to compute the first part,
$\sum_{m=1}^M\|\mathbf{q} - \mathbf{c}_{mk_m}\|_2^2$,
to find the nearest neighbors
as the second part,
$(M-1)\|\mathbf{q}\|_2^2$,
is the same
for all the database vectors.
Thus, it can be concluded
that $(M-1)$ addition operations are enough,
if we have precomputed the distance table
from the query to dictionary items
as PQ and Cartesian $k$-means do.

{\small
\bibliographystyle{splncs}
\bibliography{bow}

\begin{thebibliography}{10}

\bibitem{SameFoun2006}
Samet, H.:
\newblock Foundations of multidimensional and metric data structures.
\newblock Elsevier, Amsterdam (2006)

\bibitem{ShakhnarovichDI06}
Shakhnarovich, G., Darrell, T., Indyk, P.:
\newblock Nearest-Neighbor Methods in Learning and Vision: Theory and Practice.
\newblock The MIT press (2006)

\bibitem{Bentley75}
Bentley, J.L.:
\newblock Multidimensional binary search trees used for associative searching.
\newblock Commun. ACM \textbf{18}(9) (1975)  509--517

\bibitem{MujaL09}
Muja, M., Lowe, D.G.:
\newblock Fast approximate nearest neighbors with automatic algorithm
  configuration.
\newblock In: VISSAPP (1). (2009)  331--340

\bibitem{DatarIIM04}
Datar, M., Immorlica, N., Indyk, P., Mirrokni, V.S.:
\newblock Locality-sensitive hashing scheme based on p-stable distributions.
\newblock In: Symposium on Computational Geometry. (2004)  253--262

\bibitem{JegouDS11}
J{\'e}gou, H., Douze, M., Schmid, C.:
\newblock Product quantization for nearest neighbor search.
\newblock IEEE Trans. Pattern Anal. Mach. Intell. \textbf{33}(1) (2011)
  117--128

\bibitem{Andoni09}
Andoni, A.:
\newblock Nearest Neighbor Search: the Old, the New, and the Impossible.
\newblock PhD thesis, MIT (2009)

\bibitem{GrangierB08}
Grangier, D., Bengio, S.:
\newblock A discriminative kernel-based approach to rank images from text
  queries.
\newblock IEEE Trans. Pattern Anal. Mach. Intell. \textbf{30}(8) (2008)
  1371--1384

\bibitem{DeanRSSVY13}
Dean, T., Ruzon, M., Segal, M., Shlens, J., Vijayanarasimhan, S., Yagnik, J.:
\newblock Fast, accurate detection of 100,000 object classes on a single
  machine.
\newblock In: CVPR. (2013)  1814--1821

\bibitem{KorenBV09}
Koren, Y., Bell, R.M., Volinsky, C.:
\newblock Matrix factorization techniques for recommender systems.
\newblock IEEE Computer \textbf{42}(8) (2009)  30--37

\bibitem{BayardoMS07}
Bayardo, R.J., Ma, Y., Srikant, R.:
\newblock Scaling up all pairs similarity search.
\newblock In: WWW. (2007)  131--140

\bibitem{DeerwesterDLFH90}
Deerwester, S.C., Dumais, S.T., Landauer, T.K., Furnas, G.W., Harshman, R.A.:
\newblock Indexing by latent semantic analysis.
\newblock JASIS \textbf{41}(6) (1990)  391--407

\bibitem{GraumanD04}
Grauman, K., Darrell, T.:
\newblock Fast contour matching using approximate earth mover's distance.
\newblock In: CVPR (1). (2004)  220--227

\bibitem{RamG12}
Ram, P., Gray, A.G.:
\newblock Maximum inner-product search using cone trees.
\newblock In: KDD. (2012)  931--939

\bibitem{WeissTF08}
Weiss, Y., Torralba, A.B., Fergus, R.:
\newblock Spectral hashing.
\newblock In: NIPS. (2008)  1753--1760

\bibitem{GongL11}
Gong, Y., Lazebnik, S.:
\newblock Iterative quantization: A procrustean approach to learning binary
  codes.
\newblock In: CVPR. (2011)  817--824

\bibitem{NorouziF13}
Norouzi, M., Fleet, D.J.:
\newblock Cartesian k-means.
\newblock In: CVPR. (2013)  3017--3024

\bibitem{JainVG10}
Jain, P., Vijayanarasimhan, S., Grauman, K.:
\newblock Hashing hyperplane queries to near points with applications to
  large-scale active learning.
\newblock In: NIPS. (2010)  928--936

\bibitem{LiuWMKC12}
Liu, W., Wang, J., Mu, Y., Kumar, S., Chang, S.F.:
\newblock Compact hyperplane hashing with bilinear functions.
\newblock In: ICML. (2012)

\bibitem{MuWC12}
Mu, Y., Wright, J., Chang, S.F.:
\newblock Accelerated large scale optimization by concomitant hashing.
\newblock In: ECCV (1). (2012)  414--427

\bibitem{BasriHZ11}
Basri, R., Hassner, T., Zelnik-Manor, L.:
\newblock Approximate nearest subspace search.
\newblock IEEE Trans. Pattern Anal. Mach. Intell. \textbf{33}(2) (2011)
  266--278

\bibitem{MairalBPS09}
Mairal, J., Bach, F., Ponce, J., Sapiro, G.:
\newblock Online dictionary learning for sparse coding.
\newblock In: ICML. (2009) ~87

\bibitem{ChavezFN08}
Ch{\'a}vez, E., Figueroa, K., Navarro, G.:
\newblock Effective proximity retrieval by ordering permutations.
\newblock IEEE Trans. Pattern Anal. Mach. Intell. \textbf{30}(9) (2008)
  1647--1658

\bibitem{YuanL06}
Yuan, M., Lin, Y.:
\newblock Model selection and estimation in regression with grouped variables.
\newblock Journal of the Royal Statistical Society, Series B \textbf{68} (2006)
   49--67

\bibitem{JegouTDA11}
Jegou, H., Tavenard, R., Douze, M., Amsaleg, L.:
\newblock Searching in one billion vectors: Re-rank with source coding.
\newblock In: ICASSP. (2011)  861--864

\bibitem{JegouDS08}
Jegou, H., Douze, M., Schmid, C.:
\newblock Hamming embedding and weak geometric consistency for large scale
  image search.
\newblock In: ECCV (1). (2008)  304--317

\bibitem{BennettL07}
Bennett, J., Lanning, S.:
\newblock The netflix prize.
\newblock In: KDD Cup and Workshop. (2007)

\bibitem{GeHK013}
Ge, T., He, K., Ke, Q., Sun, J.:
\newblock Optimized product quantization for approximate nearest neighbor
  search.
\newblock In: CVPR. (2013)  2946--2953

\bibitem{DengSBL11}
Deng, J., Satheesh, S., Berg, A.C., Li, F.F.:
\newblock Fast and balanced: Efficient label tree learning for large scale
  object recognition.
\newblock In: NIPS. (2011)  567--575

\bibitem{PerronninAHS12}
Perronnin, F., Akata, Z., Harchaoui, Z., Schmid, C.:
\newblock Towards good practice in large-scale learning for image
  classification.
\newblock In: CVPR. (2012)  3482--3489

\bibitem{SanchezP11}
S{\'a}nchez, J., Perronnin, F.:
\newblock High-dimensional signature compression for large-scale image
  classification.
\newblock In: CVPR. (2011)  1665--1672

\bibitem{NisterS06}
Nist{\'e}r, D., Stew{\'e}nius, H.:
\newblock Scalable recognition with a vocabulary tree.
\newblock In: CVPR (2). (2006)  2161--2168

\end{thebibliography}
}
\end{document}